\documentclass[journal]{IEEEtran}     % Use this line for a4 paper

\IEEEoverridecommandlockouts                              % This command is only needed if 
\usepackage{amsmath}
\usepackage{cite}
\usepackage{svg}
\usepackage{graphicx}
\usepackage{algorithm,algpseudocode}
\usepackage{caption, subcaption}
\usepackage{multirow}
\usepackage{array}
\usepackage{hyperref} 
\usepackage{amssymb}
\usepackage{mathtools}
\usepackage{makecell}
\usepackage{fancyhdr}

\DeclareMathOperator*{\argmaxA}{arg\,max}
\newcommand{\rotator}{\rotatebox[origin=c]{90}}
\newcolumntype{C}[1]{>{\centering\arraybackslash}m{#1}}
\newcommand{\probP}{\text{I\kern-0.15em P}}

\title{Enhancing Object Search in Indoor Spaces via Personalized Object-factored Ontologies}

\author{Akash Chikhalikar, Ankit A. Ravankar, Jose Victorio Salazar Luces and Yasuhisa Hirata  %$^{2}$% <-this % stops a space

\thanks{This work was partially supported by JST Moonshot R\&D [Grant Number JPMJMS2034], JSPS Kakenhi [Grant Number JP21K14115], and JST SPRING [Grant Number JPMJSP2114]. (\textit{Corresponding Author: Akash Chikhalikar.)}}% <-this % stops a space
\thanks{ The authors are with the Department of Robotics, Graduate School of Engineering, Tohoku University, Japan
(a.k.chikhalikar, ankit, j.salazar, hirata@srd.mech.tohoku.ac.jp).}}%

\begin{document}
\maketitle
\thispagestyle{fancy}

\begin{abstract}
Personalization is critical for the advancement of service robots. Robots need to develop tailored understandings of the environments they are put in. Moreover, they need to be aware of changes in the environment to facilitate long-term deployment. Long-term understanding as well as personalization is necessary to execute complex tasks like \textit{prepare dinner table} or \textit{tidy my room}. A precursor to such tasks is that of \textit{Object Search}. Consequently, this paper focuses on locating and searching multiple objects in indoor environments. In this paper, we propose two crucial novelties. Firstly, we propose a novel framework that can enable robots to deduce \textit{Personalized Ontologies} of indoor environments. Our framework consists of a personalization schema that enables the robot to tune its understanding of ontologies. Secondly, we propose an \textit{Adaptive Inferencing} strategy. We integrate \textit{Dynamic Belief Updates} into our approach which improves performance in multi-object search tasks. The cumulative effect of personalization and adaptive inferencing is an improved capability in long-term object search. This framework is implemented on top of a multi-layered semantic map. We conduct experiments in real environments and compare our results against various state-of-the-art (SOTA) methods to demonstrate the effectiveness of our approach. Additionally, we show that personalization can act as a catalyst to enhance the performance of SOTAs. Video Link: \url{https://bit.ly/3WHk9i9}

\end{abstract}

\begin{IEEEkeywords}
Personalization, Probabilistic Inferencing, Object Search, Service Robotics.
\end{IEEEkeywords}

\section{Introduction}

\IEEEPARstart{P}{ersonalization} is essential for facilitating multiple long-horizon tasks. Service robots, in indoor environments, commanded with \textit{tidy up} tasks like `tidy up my living room' need to be aware of which objects are in place or out of place. Information of such nature will be specific to each house. %Additionally, they also need to act such that objects that are out of place are put in their designated places. 
Another long-horizon task belongs to the \textit{prepare} category, wherein robots can be asked to `prepare table for dinner' or `prepare bag for gym'. Humans, who perform such tasks daily, subconsciously optimize their performance using memory, foresight and holistic decision-making. In order to assimilate into human-centric environments, robots will need a personalized understanding of user preferences and environments. The concept of personalization, however, goes beyond these task execution capabilities. In the future, robots will develop personalized relationships with their owners by providing enjoyment, companionship as well as care. Apart from this, robots equipped with personalization capabilities can also help individuals with dementia. They can provide `long-term tracking' of daily activities to help individuals maintain their routines.

%Semantic Maps have a wide variety of applications such as social navigation and object search among others. 

\begingroup
\setlength{\belowcaptionskip}{-10pt}
\begin{figure}[t]
     \centering
     
     \includegraphics[width=0.44\textwidth]{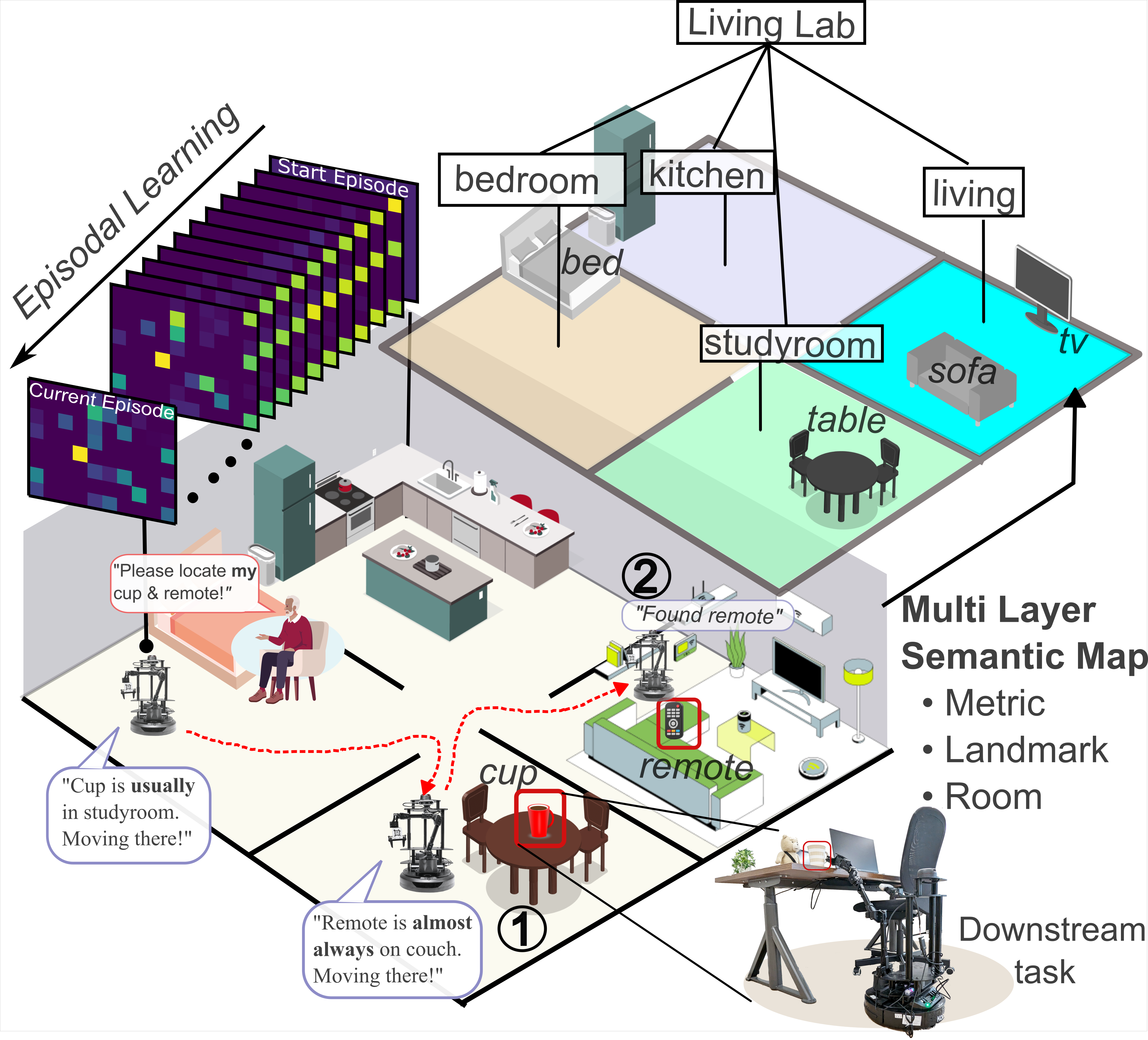}
     \caption{Our personalized object search framework. %The robot tunes its understanding of the environment over multiple episodes. 
     The robot uses adaptive inferencing and dynamic belief updates in object search.} 
     \label{fig:personalization_overview}
 \end{figure}
\endgroup

Object search tasks are primarily carried out in indoor environments. A typical indoor environment, apart from humans, consists of different rooms, various pieces of furniture, electronics (e.g., TVs, fridge), smaller items (e.g., tableware, phones) and so on. We categorize the larger, slightly immovable objects as `landmarks'. Our environment is thus a network of `object-landmark-rooms'. Searching for landmarks is a simple task due to their fixed position in the long-term. On the other hand, searching for smaller, daily-use items is a challenging task due to their dynamic and uncertain positions. These items are the `objects' for our search task. Object search is crucial for high-level downstream tasks like \textit{tidy up} (Fig. \ref{fig:personalization_overview}).

For an effective object search, the core components are the inferencing mechanism and the robot's internal belief. A convenient way to represent this belief is via an \textit{Ontology}. Ontologies are a formal representation of concepts (e.g., rooms, landmarks) within a domain (house), including the relationships between them (e.g., in, on, and near). An instance of such ontology would be: `laptop-on-table-in-living room'. Ontologies have been used for efficient representations of indoor environments \cite{vimantic}. These ontologies would be different for every environment and hence require a personalized understanding. On the other hand, the inferencing strategy should consider trade-offs between minimizing search distance and maximizing search probability. The mechanism needs to adapt to its current observations and prior beliefs and accordingly generate a plan with an appropriate horizon for the robot to follow.

To this end, the main contributions of this paper include:
\begin{itemize}
    \item A novel framework for \textit{personalization} of ontologies of various object-landmark-room relationships using an episode-based learning strategy.
    \item A novel \textit{adaptive inferencing} approach that adapts the search horizon based on the ontology priors. The approach includes \textit{dynamic belief updates} that improves performance in multi-object search tasks.%balances short-horizon and long-horizon planning.   
    \item Quantitative results obtained in real environments and comparisons with various SOTAs to demonstrate significant improvement in object search tasks.
\end{itemize}

The remainder of the paper is structured as follows. We review related work and our position in the domain in Section II. Multi-layer Semantic Mapping is presented in Section III. Section IV details our novel framework for personalization of ontologies and is the crux of this submission. In Section V we discuss our inferencing strategy and dynamic belief updates along with some experimental results. A majority of experimental results and comparisons with SOTAs can be found in Section VI. Finally, Section VII concludes the paper with a discussion on future works.
\vspace{-0.2cm}
\section{Related Work}
Object search and retrieval is a mandatory precursor task for service robots. The problem has received increased attention since the realization of maturity in robot-based object detection, semantic segmentation and indoor navigation. We review previous research in the fields of personalization, semantic mapping and object search below.
\vspace{-0.2cm}
\subsection{Personalization}
Personalization %is of deep interest to roboticists
has been explored in different domains for improving the performance of robots. Learning from user-generated pre-arranged scenes, a Graph Neural Network (GNN) is used to obtain user preference vectors and perform personalized arrangements in \cite{My_House_My_Rules}. LLMs have also shown potential as few-shot learners of personalization as demonstrated in \cite{TidyBot}. A framework for personalization of risk assessment was presented in \cite{IshiiRisk}.
Spatio-temporal models for proactive robot assistance have been discussed in \cite{pmlr-v205-patel23a}. They extended their work with an interactive query mechanism to capture human behaviour that is inherently stochastic \cite{patel2023predicting}. While spatio-temporal methods might excel in environments where object movements follow predictable patterns, they may struggle in scenarios with high variability or where objects are frequently moved to new and unseen locations. These methods also rely heavily on supervised learning and require substantial training data to model the object movement and predict future locations. %On the other hand, our method emphasizes on personalization through probabilistic inferencing based on historical observations, which allows the robot to adaptively update its belief about object locations, }
Our work, to the best of our knowledge, is a first in presenting mechanisms for deducing \textit{personalized ontologies} without explicit supervision or human robot interactions. Due to this our method can be beneficial in real-world applications where the training data may be scarce or non-existent.
\vspace{-0.2cm}
\subsection{Semantic Mapping}
Semantic Mapping is widely studied due to its foundational value. Semantic maps over long-term have been built using Truncated Signed Distance Fields (TSDFs) \cite{Panoptic_TSDFs}. Multi-robot metric-semantic mapping has been demonstrated by authors in \cite{rosinol2021kimera}. The concept of semantic exploration and object-goal navigation (ObjNav) on simulator platforms such as Habitat \cite{habitatchallenge2022} has received a lot of contributions as well. Unlike most works, which assume a pre-built semantic map for object search, we also present our framework for building a semantic map as the robot navigates through the environment.
\vspace{-0.2cm}
\subsection{Object Search} \label{Subsec:Object_Search_Related_Work}
Object search has been targeted by different researchers until now. Apart from service robotics, the problem is also interesting in the context of probabilistic inferencing and planning under uncertainty. A majority of previous works in solving the problem of object search fall into two categories: Partially Observable Markov Decision Process (POMDP) and Next Best View (NBV) methods.

\subsubsection*{\textit{POMDPs}}
Object search being a problem of decision-making under uncertainty can be inherently modelled as POMDP. The first work with this approach used belief road maps to reduce search spaces \cite{BRM_OS}.
Due to their computational intractability in large domains, a hierarchical resolution of the environment was used in \cite{Multi_scale_POMDPs}. A multi-resolution POMDP for optimizing joint rotations to maximize visual area coverage was presented in \cite{Multi-res-POMDP}. In order to optimize the POMDP-based object search, a spatial correlational model was introduced in \cite{Opt-Corr-Obj-Search}. 
\subsubsection*{\textit{NBV}}
Seemingly the most used, NBV methods treat object search as an explicit inferencing problem and different strategies are used to determine the next location. A popular approach is \cite{Viewpoint-selection}, wherein the authors determined candidate points and selected them according to belief maximization. In \cite{Semantic-Temporal}, authors prepared heat maps according to the time of day to incorporate temporal dimension in object search. On the contrary,\cite{Semantic-Grounding} proposes a temporal decay for improving search efficiency. They further improved upon their method with a heuristic that considers room sizes and distances \cite{Hierarchial_OS}. Recently, graph structures are being used for modelling the environment and planning for NBV \cite{Scene-graph-search}, \cite{Semantic_driven_OS}. NBV approaches can also be used for an open-vocabulary object search \cite{AkashOpenVocab}.  

Another approach is \cite{wang2019semantic} wherein Gaussian Mixture Models are used to prepare a costmap encoding probabilistic information and then an Informative Path Planning is carried out for search. In the absence of prior knowledge, the authors in \cite{multimodal_object_search} first determined the object category (e.g., toys, electronics) and planned robot trajectories using prior knowledge about the category.

Our work can be broadly categorized as an NBV approach. In this paper, we present a first attempt at personalization of ontologies to significantly improve performance in object search tasks. Additionally, we propose an adaptive inferencing approach for balancing long and short-horizon planning with dynamic belief updates for multi-object search. In the domain of object search, we summarize our contribution with respect to the state of the arts in Table \ref{tab:lit_compare} below.%We summarize our contribution with \textcolor{black}{respect to} some of the State of the Art in Table \ref{tab:lit_compare} below.

\begin{table}[h]
    \renewcommand{\arraystretch}{1.1}
    \centering
    \scriptsize
    \begin{tabular}{|c|c c|c c c| c c c|}
        \hline
         & \multicolumn{2}{c|}{\textcolor{black}{Scope}} & \multicolumn{3}{c|}{\textcolor{black}{Ontologies}} & \multicolumn{3}{c|}{Inferencing} \\
         \cline{2-9}
        \rule{0pt}{29pt}
        Reference & \rotator{\textcolor{black}{Single object}} & \rotator{Multi-object} & \rotator{Object-Landmark} & \rotator{Object-} \rotator{Landmark-Room} & \rotator{\textcolor{black}{Personalized}} & \rotator{Dynamic} \rotator{Belief Update} & \rotator{\textcolor{black}{Fixed-horizon}} & \rotator{\textcolor{black}{Adaptive}}  
        \\[0.8cm]
        \hline
        \cite{Opt-Corr-Obj-Search} & \checkmark & - & \checkmark & -  & - & \checkmark & \checkmark &- \\
        %\hline
        \cite{Viewpoint-selection} & \checkmark & - & \checkmark & -  & - & \checkmark & \checkmark &- \\
        %\hline
        \cite{Semantic-Temporal} & \checkmark & - & \checkmark & -  & - & - & \checkmark &- \\
        %\hline
        \cite{Semantic-Grounding}%$^{\#}$
        & \checkmark & \checkmark & \checkmark & \checkmark & - & \checkmark & \checkmark &- \\
        %\hline
        \cite{Scene-graph-search}%$^{\#}$
        & \checkmark & \checkmark & \checkmark & - & - & \checkmark & \checkmark & - \\
        %\hline
        Ours &  \checkmark & \checkmark &  \checkmark & \checkmark & \checkmark & \checkmark & - & \checkmark \\
        \hline
        %\multicolumn{5}{l}{\scriptsize *LTC = Long-term Capabilities}
    \end{tabular}
    \captionsetup{belowskip=-15pt}
    \caption{Positioning with respect to related works of object search}    
    \label{tab:lit_compare}
\end{table}

\section{Multi-layer Semantic Mapping}
Object search is a complex task in human-centric environments. %It is a consequence of higher-level cognitive search and lower-level navigation capabilities.
\textcolor{black}{Semantic Map functions as an important preliminary for object search. The type of semantic map the robot has access to, will significantly influence the level of personalization the robot can achieve, and the sequence of locations it would visit when assigned a task.} This section briefly describes each layer of our semantic map and how it was generated.%Unlike many researchers who either use digital simulators or assume a pre-built semantic map, we prepare our semantic map by navigating the robot through the environment.

\subsubsection*{Metric Layer} \label{SubSec:Mapping}
A 2-dimensional occupancy grid layer for which, we fuse data from RGB-D sensor, 2D Lidar and wheel odometry \textcolor{black}{using visual SLAM techniques}. %to create a dense 3D point cloud map (octomap) of the environment. 

\subsubsection*{Semantic Layer}
The semantic layer consists of ubiquitous landmarks such as sofa, bed, etc. \textcolor{black}{We use YOLOv7 \cite{yolov7} for object detection.} The corresponding point cloud data is randomly sampled, and the mean coordinates are computed \textcolor{black}{for localization of the landmark on the map}.%During the mapping process, YOLOv7 \cite{yolov7} is employed to detect objects in frames. 
%The YOLOv7 network generates a 5-D output for each detected object, including the object class and four bounding box parameters (centre coordinates: $C_x$ and $C_y$, breadth: $B$, and length: $L$). 

%The bounding boxes are randomly sampled and the mean $X$ and $Y$ coordinates are computed from the corresponding point cloud data \textcolor{black}{for localization of the landmark on the map}. Using average pooling, multiple measurements are combined .

After localization, persistent tracking of landmarks is necessary to distinguish between new and prior observations of the objects. For such persistent tracking, we maintain a record of $K$ prior observations as $\textbf{P}=\{P_0,\cdots,P_K\}$ and new observations in the latest frame as $\textbf{C}=\{C_0,\cdots,C_L\}$. \textcolor{black}{We compute an association matrix ($\textbf{D}_{k,l}$) using Euclidean distance as shown below:} 
%We then use Euclidean distance to calculate an association matrix $\textbf{D}_{k,l}$ between both the sets as shown in equation \ref{eq:Euclidean Distance} below. 
%\uvec{D}_{k,l}

\begingroup
\fontsize{9.5}{1.5}\selectfont
\begin{align}
\label{eq:Euclidean Distance}
    \forall (k,l) \in (K,L):\ \textbf{D}_{k,l} = \sqrt{(\textbf{P}_k-\textbf{C}_l)^T(\textbf{P}_k-\textbf{C}_l)}
\end{align} 
\endgroup

After the association matrix is computed, the association between new observations (i.e., set $\textbf{C}$) and prior observations (i.e., set $\textbf{P}$) is determined using the Hungarian algorithm\cite{kuhn1955hungarian}. A Kalman
filter is then used to localize the landmark based on the associated new
observation. For more details related to this implementation, please refer \cite{akashSII}.

\vspace{0.1cm}

\subsubsection*{Segmentation and Classification Layer} \label{Subsec:Seg+Classify}
Additional information regarding room regions is necessary for inferencing based on object-landmark-room ontologies. To achieve this, we segment the metric map using Voronoi-graphs \cite{ipa_room_segmentation}, and then assign room labels to each region based on empirical data. In a segmented region $R$, containing a set of landmarks \textbf{L}, with $\|\textbf{L}\|=n$, the room label is computed as:

 \begingroup
 \fontsize{9.5}{1.5}\selectfont
\begin{align}
    \label{eq:Room_Labeling}
    \begin{aligned}
        &P(R_{j}|L_{i}) = \frac{F_{ij}}{\sum\limits_{j=1}^{N_{r}}F_{ij}}\\
        &P(R_{j}|L_{1},...,L_{n})=\prod\limits_{i=1}^{n}P(R_{j}|L_{i})
    \end{aligned}
\end{align}
 \endgroup

\begin{figure}[h!]
    \centering
   \includegraphics[width=0.45\textwidth]{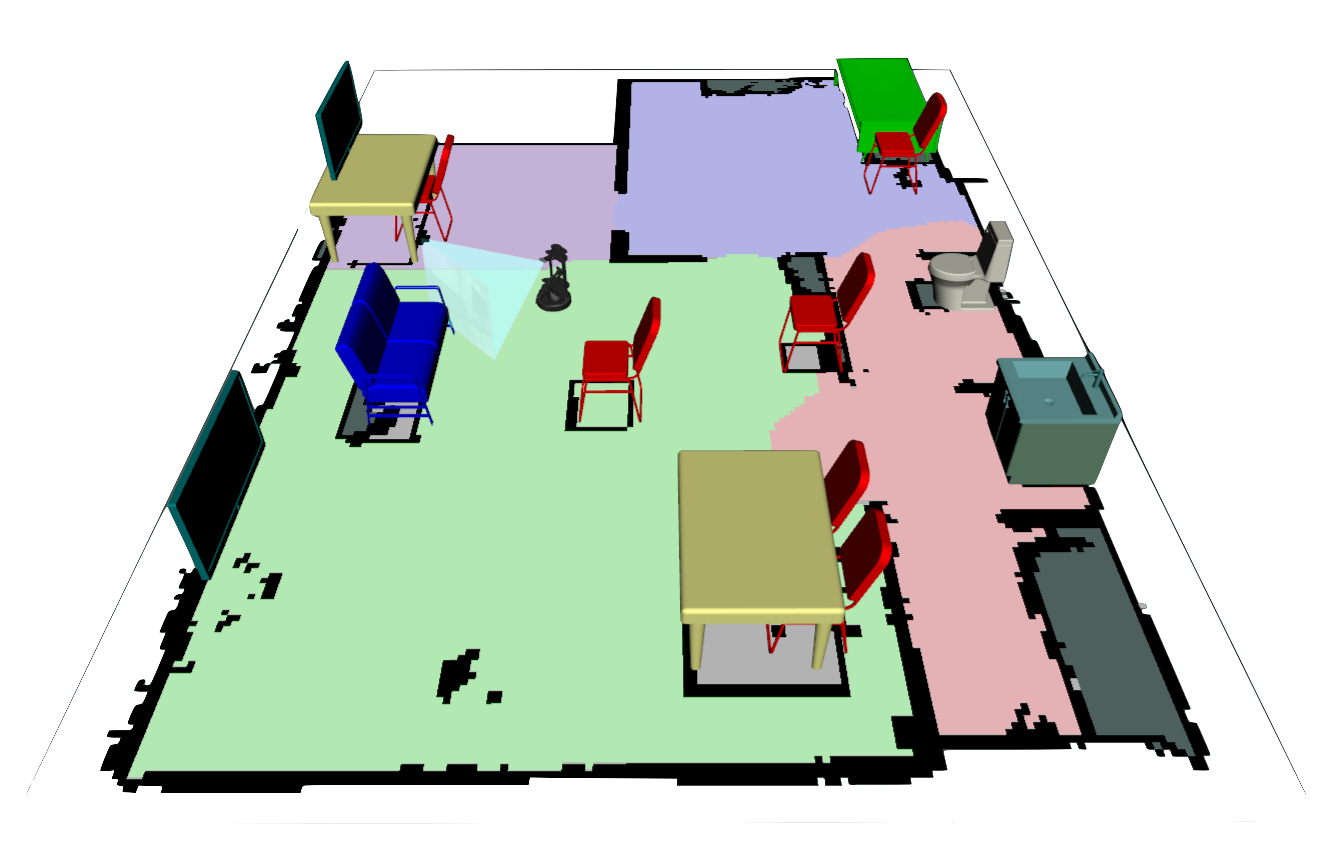}
   \captionsetup{belowskip=-8pt,aboveskip=-1pt}
    \caption{Multi-layered Semantic Map}
    \label{fig:Semantic-Map}
\end{figure}

\textcolor{black}{W}here, $F_{ij}$ accounts for occurrences of the object category `i' in the room category `j' in the Places365 \cite{Places365} dataset. The number of room categories $N_{r}$ is set to 4 [Bedroom, Living room, Bathroom, Study room]. Finally, a Multi-layered Semantic Map is obtained (Fig. \ref{fig:Semantic-Map}).

%\vspace{-0.3cm}

\section{Personalization framework}\label{Sec: Personalization_Framework}
We define personalization as the strategy by which a robot can understand preferences of a user pertaining to the object locations within an environment. We interpret the preferences as `personalized ontologies'.
%n this section, we present our framework for personalization of ontologies that will help robots get a deeper understanding of their surroundings. 
For example, a general ontology would be: cups-on-tables. However, if there are multiple tables in different rooms, such ontology wouldn't be informative. If a user spends more time in the living room than in the study room, then ontologies should prioritize the presence of a cup on the living room table over the study room table. 
In principle, a landmark in the study room would be weighted differently from a landmark in the living room. Hence, a personalized ontology would have different weights for `cup-on-table-in-living room' vs `cup-on-table-in-dining room'.
%In a more organised house, for example, the TV remote is more likely to be found in the TV's vicinity. On the other hand, in some houses, the infamous couch may be the most cluttered piece of furniture present. 
It is precisely these weights that we intend our robot to learn. % in the form of probabilistic relationships 
 %We sample the locations of objects for an episode based on a multinomial distribution 
Our strategy combines adaptive episode-based belief updates with a reliable termination criterion to obtain this personalized knowledge.

%\vspace{-0.3cm}

\subsection{Ontology Personalization Schema}
\begin{figure*}[t!]
    \centering
    \begin{subfigure}[b]{0.3\textwidth}
        \centering
        \includegraphics[width=\textwidth]{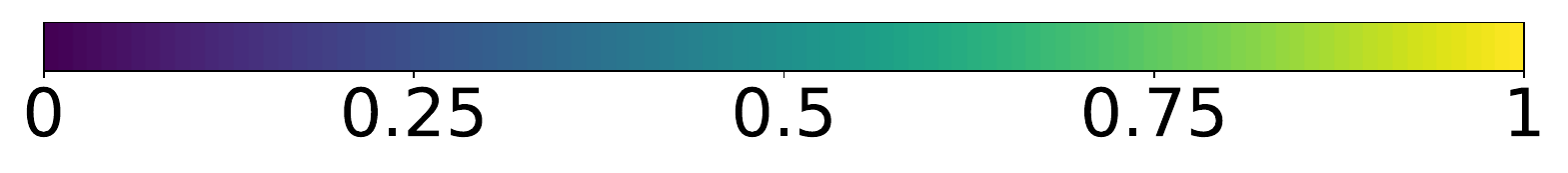}
        \label{Fig:Colorscale}
    \end{subfigure}
    
    \vspace{-0.35cm}
    
    %\rotatebox[origin=lB]{90}{\parbox{4.5cm}{\centering \scriptsize Search Objects}}
    \begin{subfigure}[c]{0.245\textwidth}
    \centering
        \includegraphics[width=\textwidth]{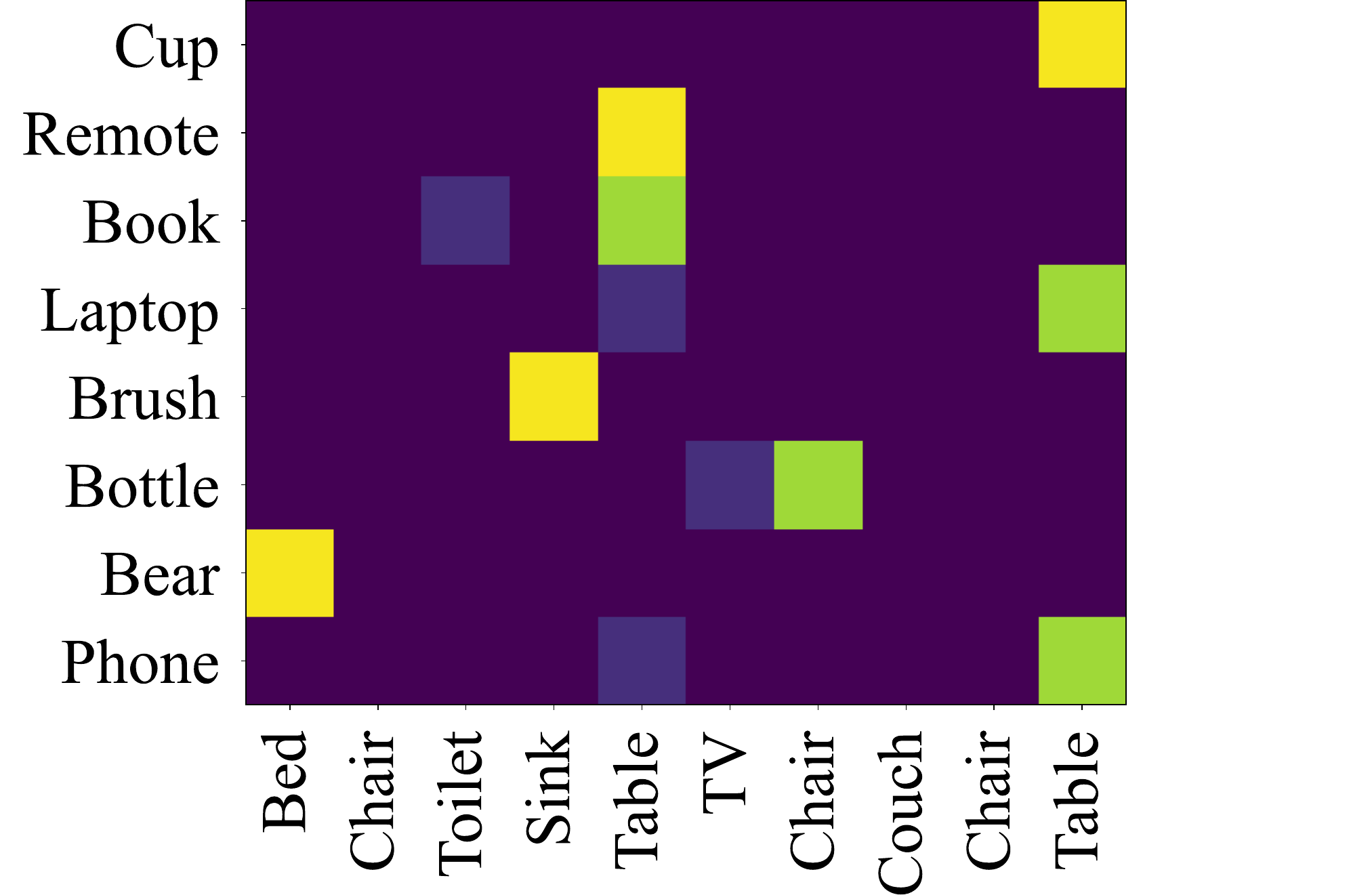}
        \captionsetup{justification=centering}
        \caption{After episode 1}
        \label{Fig:After_Episode_0}
    \end{subfigure}
    \hfill
    \begin{subfigure}[c]{0.245\textwidth}
        \centering
        \includegraphics[width=\textwidth]{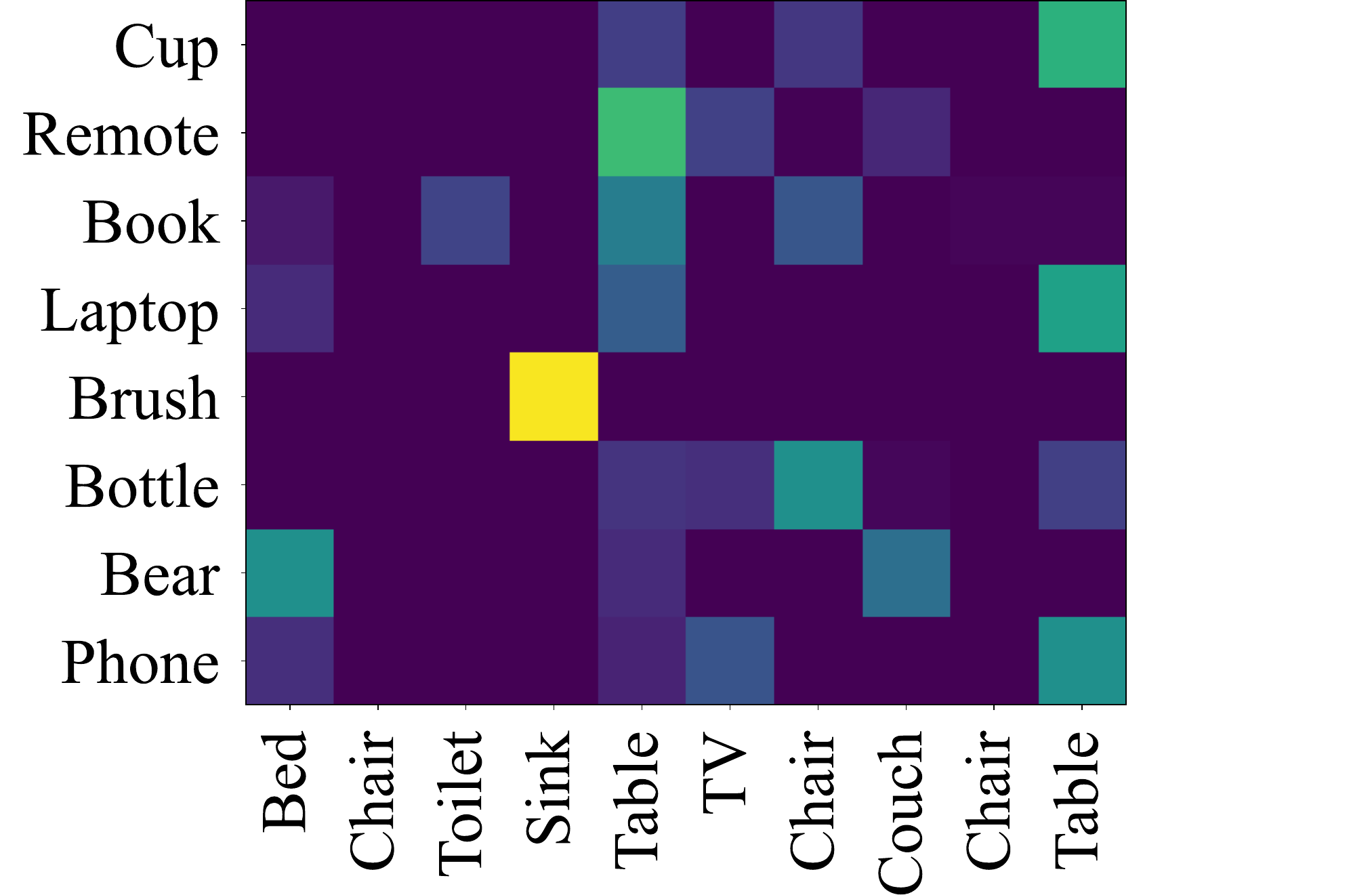}
        \captionsetup{justification=centering}
        \caption{After episode 30}
        \label{Fig:After_Episode_30}
    \end{subfigure}
    \hfill
    \begin{subfigure}[c]{0.245\textwidth}
        \centering
        \includegraphics[width=\textwidth]{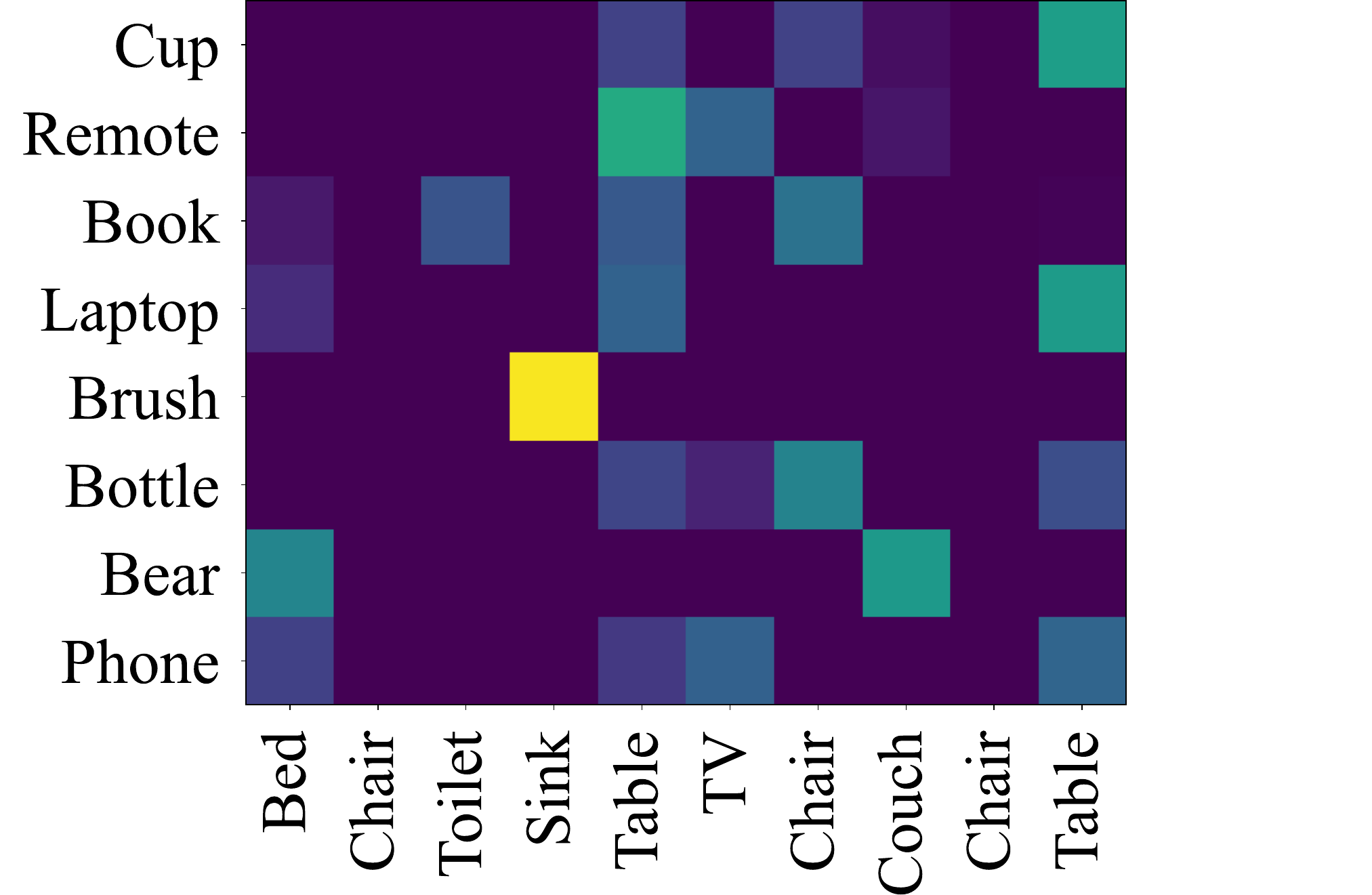}
        \captionsetup{justification=centering}
        \caption{After episode 62}
        \label{Fig:After_Episode_61}
    \end{subfigure}
    \hfill
    \begin{subfigure}[c]{0.245\textwidth}
        \includegraphics[width=\textwidth]{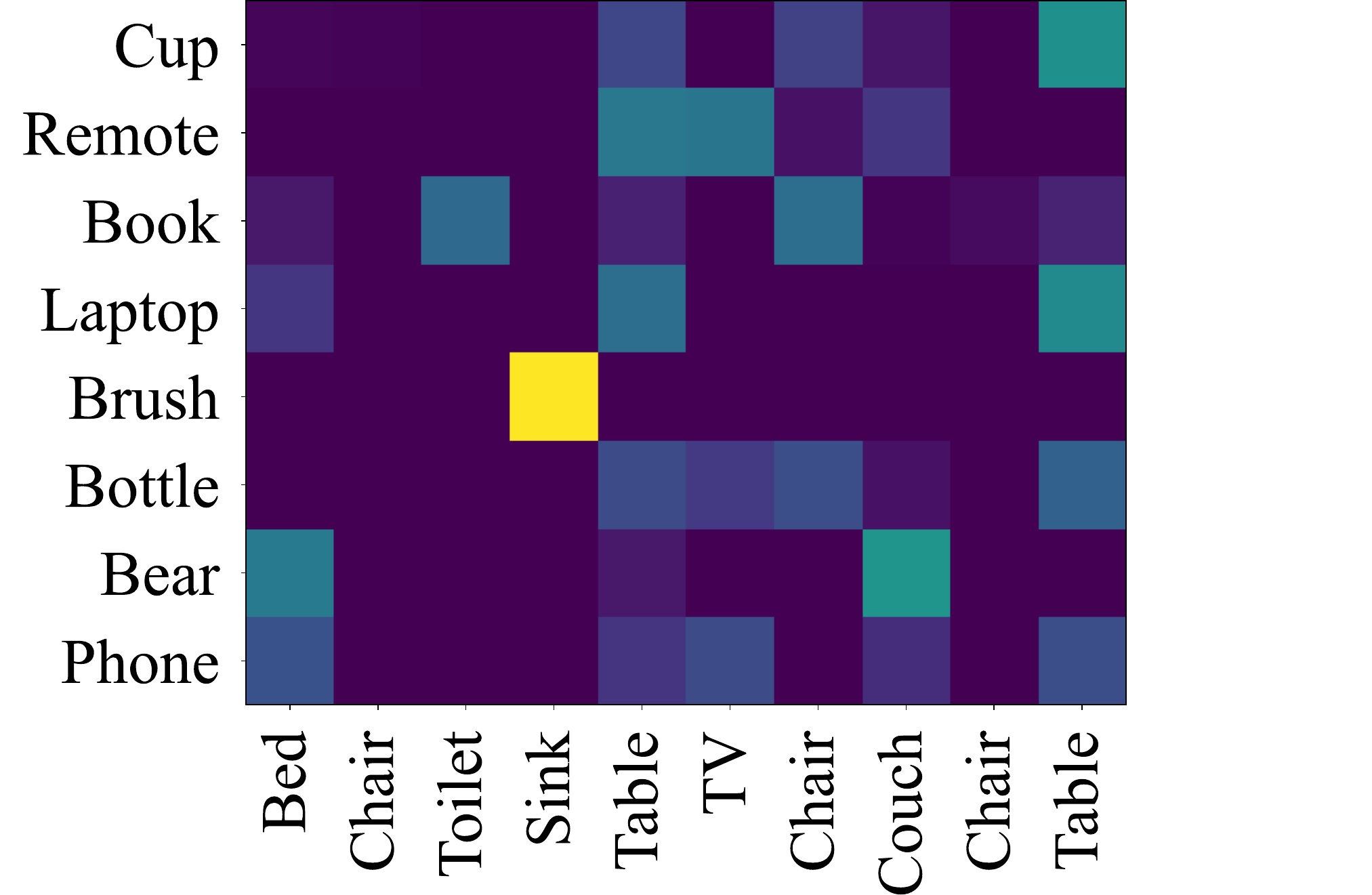}
        \captionsetup{justification=centering}
        \caption{True values}
        \label{Fig:True Values}
    \end{subfigure}

    %\vspace{0.1cm}
    
    %\rotatebox[origin=lB]{90}{\parbox{3.7cm}{\centering \scriptsize Object: Phone}}
    %\hspace{0.1cm}
    \begin{subfigure}[b]{0.24\textwidth}
        \includegraphics[width=\textwidth]{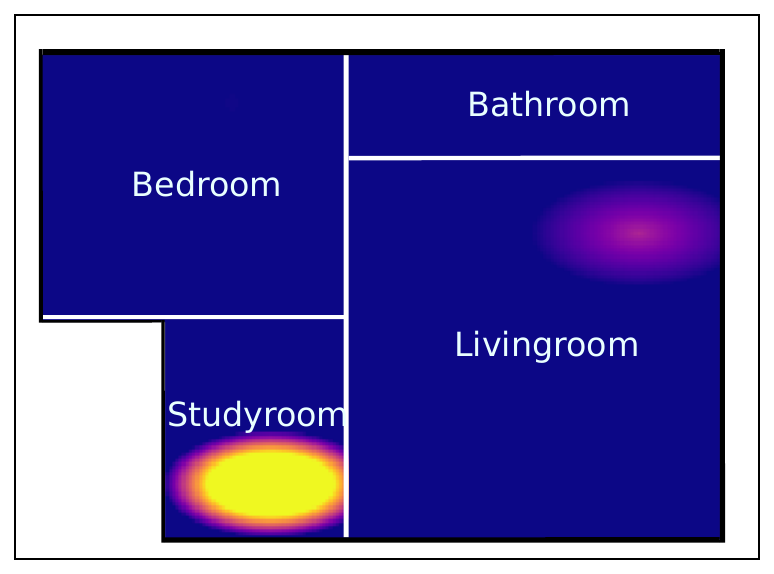}
        \captionsetup{justification=centering}
        \caption{After episode 1}
        \label{Fig:Heatmap_After_0}
    \end{subfigure}
    \hfill
    \begin{subfigure}[b]{0.24\textwidth}
        \includegraphics[width=\textwidth]{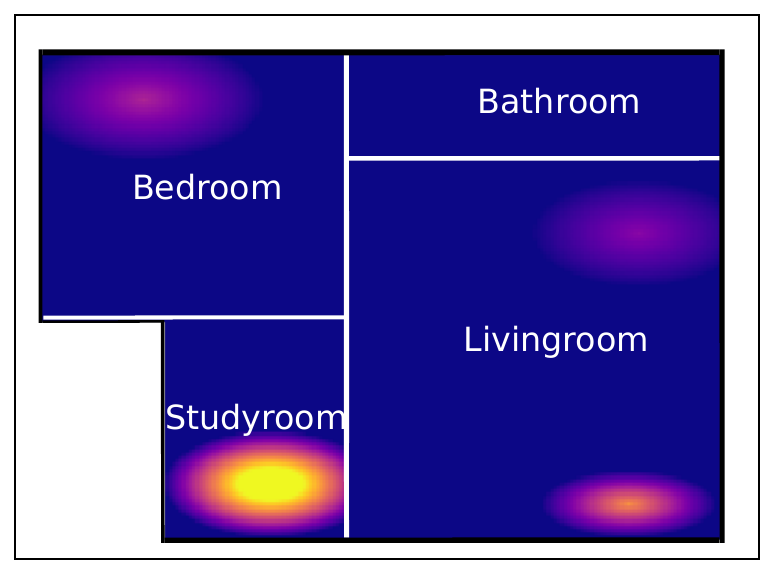}
        \captionsetup{justification=centering}
        \caption{After episode 30}
        \label{Fig:Heatmap_After_30}
    \end{subfigure}
    \hfill
    \begin{subfigure}[b]{0.24\textwidth}
        \includegraphics[width=\textwidth]{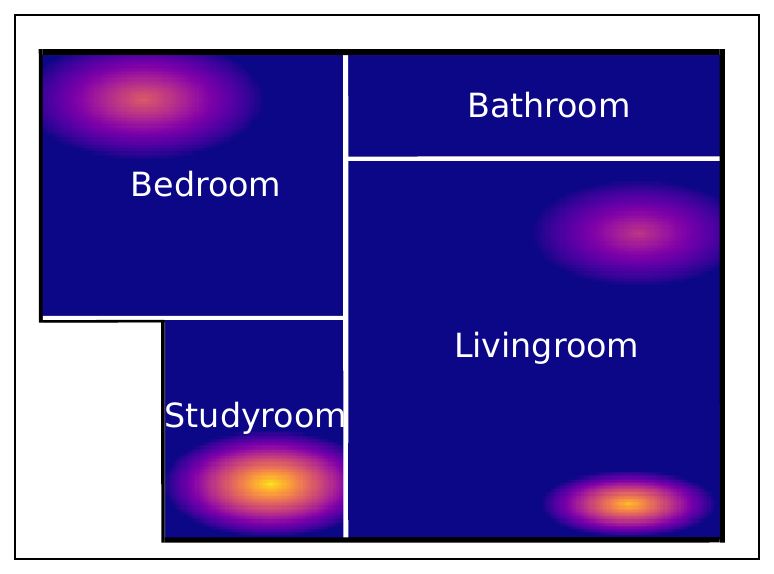}
        \captionsetup{justification=centering}
        \caption{After episode 62}
        \label{Fig:Heatmap_After_61}
    \end{subfigure}
    \hfill
    \begin{subfigure}[b]{0.24\textwidth}
        \includegraphics[width=\textwidth]{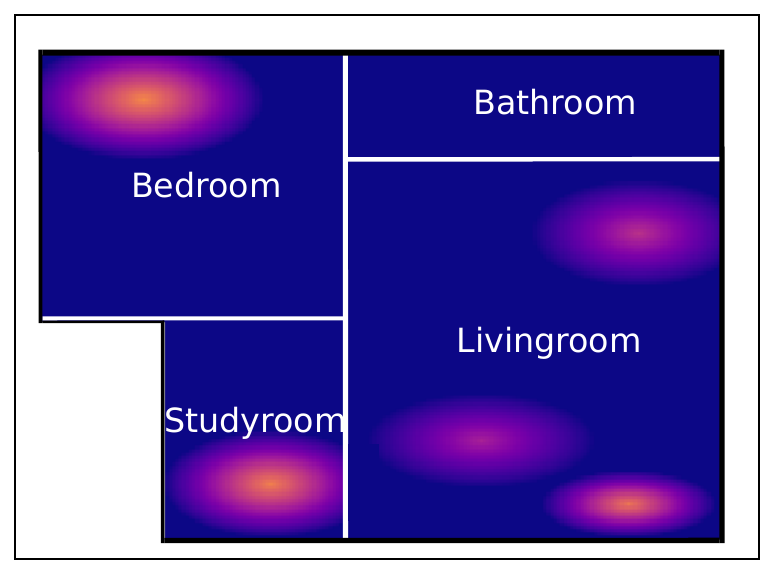}
        \captionsetup{justification=centering}
        \caption{True heatmap}
        \label{Fig:Heatmap(True)}
    \end{subfigure}
    % \vspace{-0.1cm}
    \captionsetup{belowskip=-8pt}
    \caption{In \{a-c\}, based on the uniform initial estimate, the probabilistic relationship between objects and landmarks is shown for different episodes. After every episode, the robot updates its belief (that is, $P(O|T)$), which starts resembling the true values \{d\}. \{e-g\} shows the heatmap progression for the object: Phone. The heatmap is populated based on observations (normalized) during each episode. \{h\} represents the heatmap based on the true \textit{phone-landmark-room} relationship.
    %In \{i-l\}, the different trajectories taken by the robot during the episodes are shown. The trajectories become shorter as the episodes progress.
    }
    \label{fig:Personalization Preview}
\end{figure*}

We use an episodic process to develop personalized ontologies between different objects, landmarks and rooms present in the environment. The true probabilistic relationships between object-landmarks-rooms, for these episodes, are shown in Fig. \ref{Fig:True Values}. For each episode, we sample the locations of objects based on a multinomial distribution of the true values. Then the robot navigates through the environment and updates its belief. After each episode, the locations are changed as per the sampling and the process is reiterated. In this context, the term episode refers to the entire sequence where the robot navigates through the environment once, collects observations, and updates its belief. With this setup, the robot learns the personalized probabilistic relationships.
We consider a uniform initial estimate, wherein, all objects have equal probability at all landmarks, irrespective of rooms. We represent the landmark-room tuple as \(T = (L, R)\). Later (in Sec. \ref{Sec: Initial_Estimate}), we analyze the results with different initial estimates.

After each episode, $P(O|T)$ changes (Fig. \ref{Fig:After_Episode_0}-\ref{Fig:After_Episode_61}). To infer the probabilistic relationships for the current episode, a discounted sum of the previous episodes is considered as follows:

\begingroup
\fontsize{9.5}{1.5}\selectfont
\begin{align}
    %\begin{aligned}
        P_{\tau+1}(O|T)=\displaystyle\sum\limits_{t=0}^{H}\gamma^{t}P_{\tau-t}(O|T)
    %\end{aligned}
    \label{Eq:Ontology_Prediction}
\end{align}
\endgroup

Where, $\gamma(=0.9)$ is the discounting factor and $H(=5)$ is the finite horizon. During each episode, the landmarks are ranked based on the aggregate probabilities of all objects and visited according to a probabilistic greedy strategy.

%\vspace{-0.3cm}

\subsection{Multi-resolution P-learning}
Assume that during any episode an object $O$ was observed at landmark $L$ in Room $R$ (i.e., \(T = (L, R)\)). To develop personalized ontologies, firstly, the robot needs to update the beliefs at both the room and landmark level. Secondly, updating the beliefs at all other locations ($T'$) uniformly is also not appropriate. A landmark of the same category in a different room should be updated differently from a different landmark in the same room. Hence, a multi-resolution update is more suitable than mere frequency-based updates.  Considering such conflicting conditions, we propose the equations for Multi-resolution P-learning as follows:

\begingroup
\fontsize{9.5}{1.5}\selectfont
\begin{align}
    \label{eq:Prob_update_1}
    P_{t+1}(O|T) &= P_{t}(O|T)+\frac{\beta}{\sqrt{N}}\left[1-P_{t}(O|T)\right]
\end{align}
\begin{align}
    \label{eq:Prob_update_2}
    P_{t+1}(O|T') &=P_{t}(O|T')-\frac{\beta S(T,T')}{\sqrt{N}}\left[1-P_{t}(O|T')\right]
\end{align}
\endgroup

Eq. \ref{eq:Prob_update_1} updates the probability at the observed location while Eq. \ref{eq:Prob_update_2} updates for all other unobserved locations. $S$(.) is the `similarity kernel' which helps in appropriate scaling of the probability updates at unobserved locations. It is defined as:

\begingroup
\fontsize{9.5}{9.5}\selectfont
\begin{align}
    S(T,T') = 2-\left[\delta(L,L')+\delta(R,R')\right]
\end{align}
\endgroup

Where, $\delta$(.) is the Kronecker-delta function. $\beta(=0.1)$ is the learning rate and $N$ is the episode number. The probabilities are re-normalized after each update step.

\vspace{-0.1cm}

\subsection{Confidence based termination}
While the aforementioned episodal learning strategy will converge to true values by the law of large numbers, it is necessary to establish a quicker convergence criterion.

The presence of objects at different landmarks and rooms can be modelled as a multinomial distribution. However, determining confidence interval (C.I) for multinomial distributions requires a large number of samples. In contrast, whether a room contains an object or not can be modelled as a binomial distribution. We can thus compute C.Is for all rooms and leverage the intervals to establish a termination criterion.

Since the robot learns the probabilistic relationships between objects, landmarks and rooms, we can find the converse as follows:

\begingroup
%\small
\fontsize{9.5}{1.5}\selectfont
\begin{align}
    P(R|O) = \frac{P(O|R)P(R)}{\displaystyle\sum\limits_{\forall R}P(O|R)P(R)}
\end{align}
\endgroup

The next step is to calculate the C.I for \(P(R|O)\). We use Wilson C.I as it provides better results when the distribution is skewed in either direction and is calculated as follows:
%The Wilson C.I provides better results when the distribution is skewed in either direction w.r.t Wald C.I. Additionally, it provides a leaner interval as compared to Agresti-Coull C.I for the same confidence level. The Wilson Confidence Interval is calculated as follows: 

\begingroup
\fontsize{9.5}{1.5}\selectfont
\begin{align}
    C.I = \frac{zN}{z^{2}+N}\cdot\sqrt{\sigma^2+\frac{z^{2}}{4N^{2}}}
\end{align}
\endgroup

Where, $z=1.96$ corresponds to 95\% quantile  and $\sigma$ is Standard Deviation for the binomial distribution. When the 95\% C.I is within 0.05 for all objects, we terminate the episodes.

In Fig. \ref{fig:object given room}, the progression of probability $P(O|R)$ against episode number, starting with the uniform initial estimate, 
is shown for two objects: laptop and phone. The corresponding heatmap progression for phone is given in Fig. \ref{Fig:Heatmap_After_0}-\ref{Fig:Heatmap_After_61}. In our scenario, the laptop is much more likely to be in the study room than the bedroom whereas, the phone is equally likely to be in the living or the study. Consequently, the personalization results in $P(\text{Phone}|\text{Living}) \approx P(\text{Phone}|\text{Study})$ and $P(\text{Laptop}|\text{Study}) \gg P(\text{Laptop}|\text{Bedroom})$.
%Check which is better, $ or no $. 
It is important to note that the framework is robust to any sort of personalization as per user preferences.

\begingroup
\setlength{\belowcaptionskip}{0pt}
\begin{figure}[H]
    \begin{subfigure}[]{0.24\textwidth}
        \includegraphics[width=\textwidth]{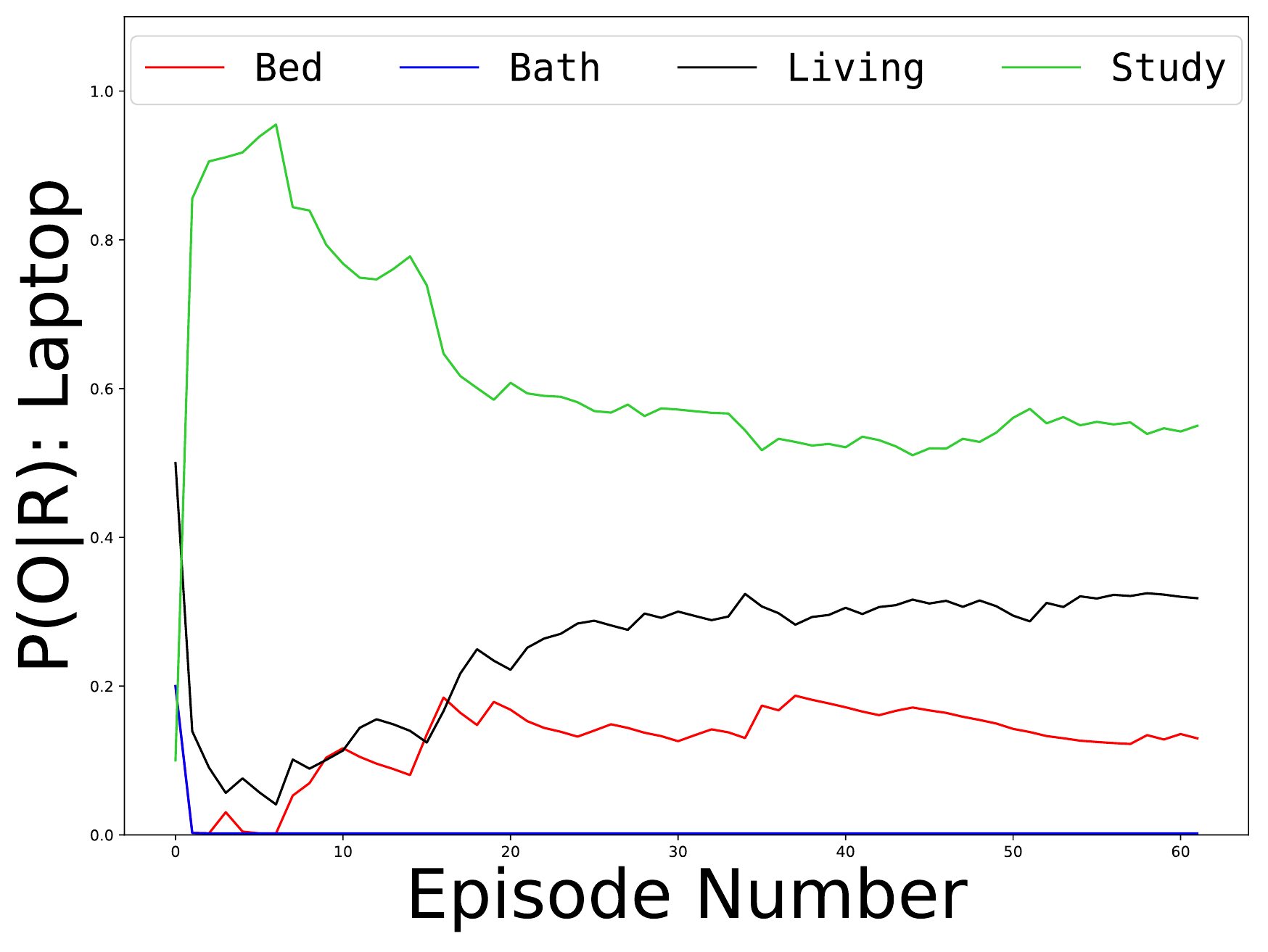}
        \captionsetup{justification=centering}
        %\caption{Target object: Book}
        \label{Fig:laptop given room}
    \end{subfigure}
    \hfill
    \begin{subfigure}[]{0.24\textwidth}
        \includegraphics[width=\textwidth]{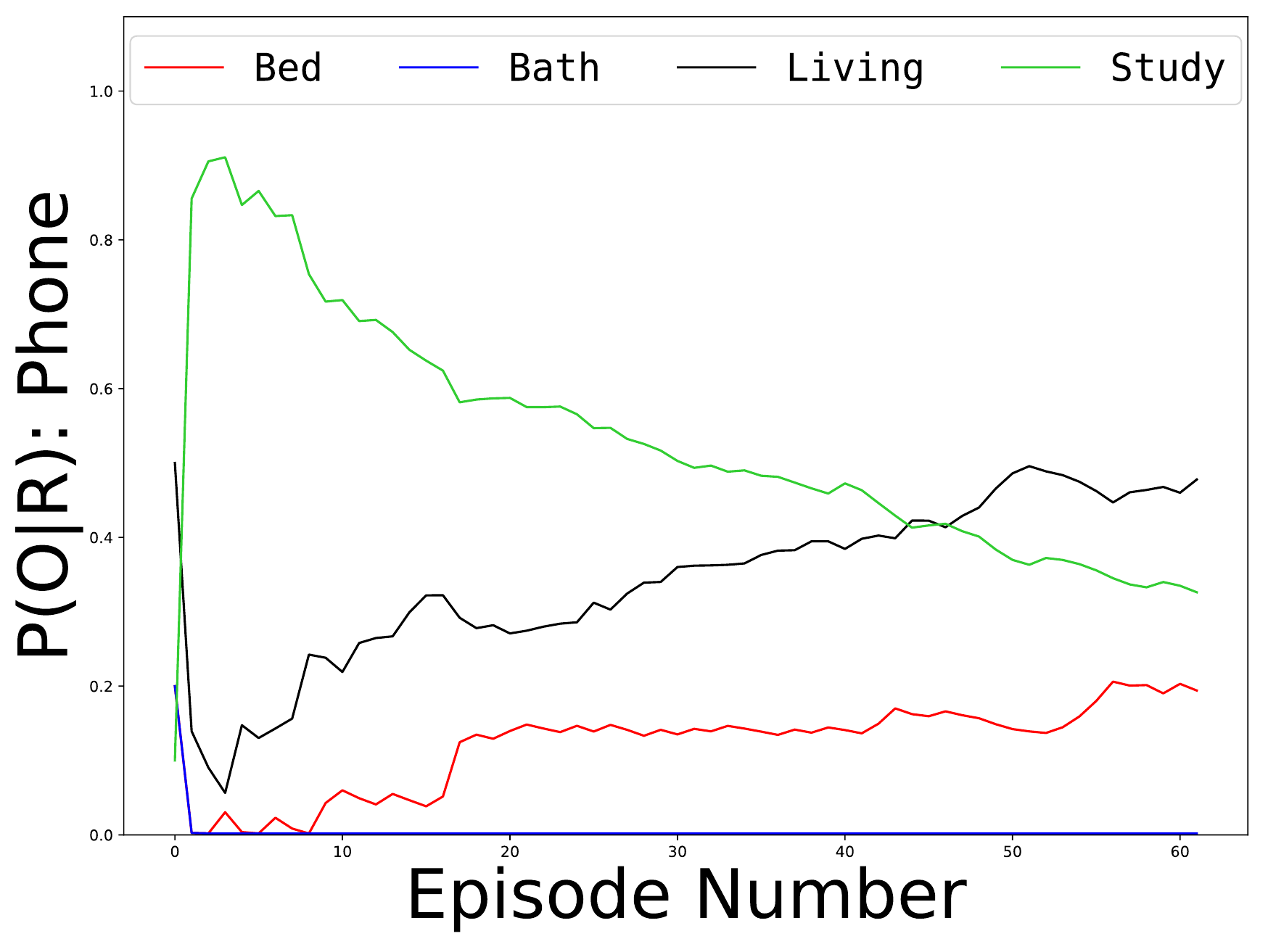}
        \captionsetup{justification=centering}
        %\caption{Target object: Phone}
        \label{Fig:phone given room}
    \end{subfigure}
    \captionsetup{justification=centering,skip=-5pt}
    \caption{Probability of object given room}
    \label{fig:object given room}
\end{figure}
\endgroup

\section{Long-term Object Search}
We propose an adaptive inferencing strategy with dynamic belief updates for enabling object search in the long-term. It is important to note that our inferencing approach is decoupled from the personalization strategy since it only depends on the ontology prior, irrespective of whether it is personalized. % or not.  %We assume from hereon that all probabilistic relationships have been personalized as a result of our proposed framework.

\vspace{-0.3cm}

\subsection{Inferencing}
The task of searching discrete locations to minimize the time to detection is NP-hard \cite{Complexity-search-path}. Hence, sub-optimal solutions using approximations or heuristic-based methods are necessary. A short-horizon strategy %such as `Probabilistic Greedy'
yields better results if the beliefs are skewed towards a particular landmark. For e.g., visiting a location $T$, where $P(O|T)>0.5$, will lead to a quicker search more than 50\% of the time, irrespective of the distances. On the other hand, for a more uniform probabilistic distribution, a long-horizon strategy that considers a trade-off between probability and distance will lead to better results. \textcolor{black}{Essentially, the strategy should adapt the search horizon according to the distribution.} Thus, we use an inferencing strategy that is a hybrid between short-horizon and long-horizon planners. The authors in \cite{Decision_making_framework} suggest an Optimal Lookahead Search: a form of receding horizon control  transposed to solve the search path-planning problem. Their method consists of a hyper-parameter `w' which is the \textit{lookahead window}. We propose a formal approach to determine this window based on the Cumulative Distribution Function (CDF) of the object (see Line 8, Algorithm \ref{Alg:Inference_Algo}). After, determining the lookahead window, we calculate the utility of each tuple in it as follows:% the window as follows:

\begingroup
\fontsize{9.5}{1.5}\selectfont
\begin{align}
    \label{eq: Heuristic}
    U_{T}\ = P(O|T)+\frac{\alpha}{Dist(T,A)} 
\end{align}
\endgroup

%\vspace{-0.3cm}

Where, $Dist(T,A)$ represents the distance between the landmark and agent (robot) obtained via the $A$-star planner and $\alpha$ is the scaling parameter. With the utilities computed, object search is transposed as a Travelling Salesperson Problem (TSP) with known start and arbitrary ends. We use Google-OR TSP solver CP-SAT to solve this problem and determine the next landmark. The process begins again if the object is not found. 
%Our strategy performs better than SOTA methods as we demonstrate in Sec. \ref{subsec:SOTA_Comparison}. This is a direct consequence of our personalization approach paired with our inferencing strategy. 
We summarize our strategy in Algorithm \ref{Alg:Inference_Algo}.

\vspace{-0.3cm}

\subsection{Dynamic Belief Update} \label{Sec:DBU}
In long-term deployments, the robot may be repeatedly tasked with searching for objects, during which the state of the environment may not change much. In such scenarios, it is necessary for the inferencing strategy to adapt to information gained while the search is ongoing. When tasked with searching for an object(s), this manifests as the capability of tracking other objects during the search. We use a dynamic belief update (DBU) mechanism to do the same.

%Algorithm for inferencing
\setlength{\textfloatsep}{8pt}
\begin{algorithm}[t]
    \caption{Algorithm for inferencing}\label{Alg:Inference_Algo}
    \begin{algorithmic}[1]
        \Require $\text{Map}:\mathcal{M},\ T=(L,R),\ \text{Belief}:\ P(O|T)$
        \State $\textbf{Input:}\ Search\ Object (O^{*})$
        \While{$Not\_Found$}
            \State $Sort[P_{t}(O^{*}|T)]$
            \If{$Max[P_{t}(O^{*}|T)] > 0.5$}
            \State $T_{next}\gets argmax_{T} P(O^{*}|T)$
            %\State $A = Visit (T)$
            \State $A = Check(T_{next}|O^{*})$
            \Else
            \State $WindowSize = Size[CDF\{P_{t}(O^{*}|T)\}>0.5]$ 
            \For {$i$ in $WindowSize$}
            \State $U_{T} \gets P_{t}(O^{*}|T)+\alpha/D(T,Agent)$
            \EndFor
            \State $T_{next} \gets TSPSolver(U_{T},X)$
            \State $A = Check(T_{next}|O^{*})$
            \EndIf
            \State $P_{t+1}(O|T)\gets DBU (P_{t}(O|T),T)$ \Comment{$\forall Objects$}
        \EndWhile
    \end{algorithmic} 
    %\vspace{-0.1cm}
\end{algorithm}

Consider that a robot has visited a set of landmarks $\textbf{L}_{s} \subset \textbf{L}$ before time $t$. Next, at time $t$, the robot approached landmark $L \in \textbf{L}-\textbf{L}_{s}$. In such cases, $\forall$ Objects ($O$), the belief for the next time step is updated as follows:

$\text{If}\ O \in \text{Found}\ at\ L,\ \ \forall\ \widetilde{L}\in \textbf{L}-\textbf{L}_{s}-\{L\} $:

\begingroup
\fontsize{9.5}{1.5}\selectfont
\begin{align}
    \begin{aligned}
    P_{t+1}(O|T) &= 1 \\
    P_{t+1}(O|\widetilde{T}) &= 0
    \end{aligned}
\end{align}
\endgroup

Else:

\begingroup
\fontsize{9.5}{1.5}\selectfont
\begin{align}
    \begin{aligned}
        P_{t+1}(O|T) &=0 \\
        P_{t+1}(O|\widetilde{T}) &= \frac{P_{t}(O|\widetilde{T})}{\left[1-P_{t}(O|T)\right]}
    \end{aligned}
\end{align}
\endgroup

\begin{figure}[h]
    \centering
    \begingroup
    %\fontsize{10.0}{1.5}\selectfont
    \begin{subfigure}[b]{0.49\textwidth}
        \includegraphics[width=\textwidth]{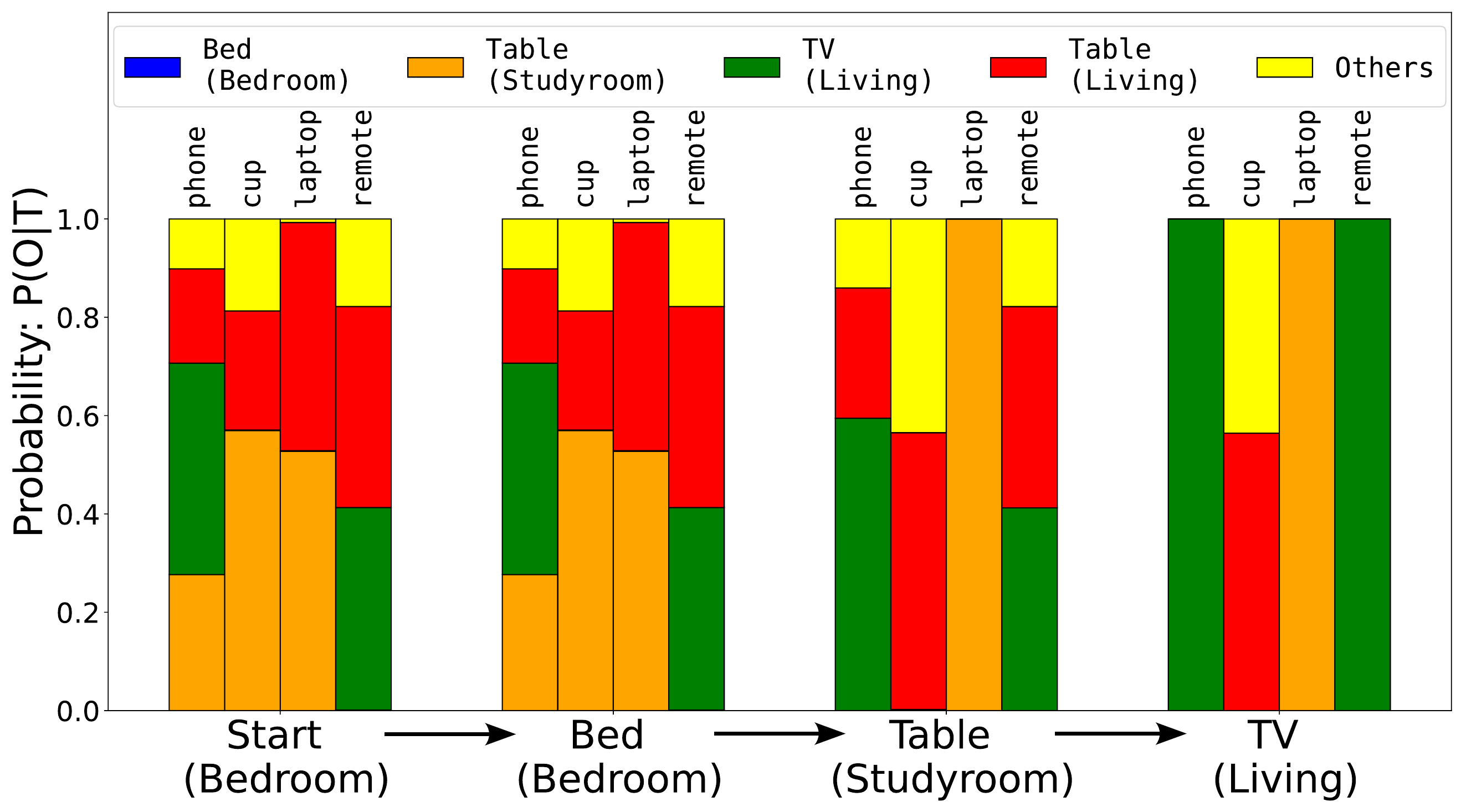}
        \captionsetup{justification=centering, skip=5pt}
        \caption{Belief propagation: P(O$|$T)}
    \end{subfigure}
    \endgroup
    
    \vspace{0.1cm}

    \subcaptionsetup{font=normalsize}
    \subcaptionbox{Heatmap propagation for object: Bottle}{
    \begin{subfigure}[b]{0.12\textwidth}
        \includegraphics[width=\textwidth]{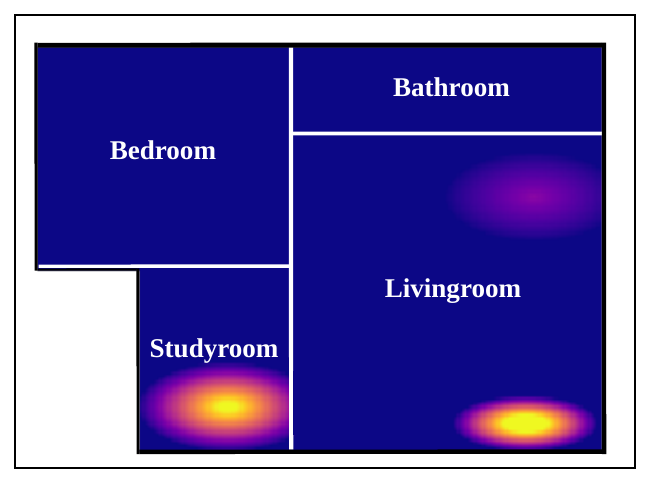}
    \end{subfigure}
    \hfill
    \begin{subfigure}[b]{0.12\textwidth}
        \includegraphics[width=\textwidth]{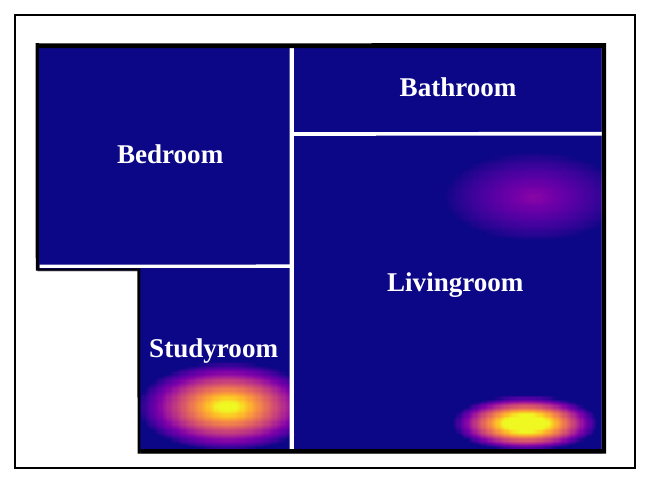}
    \end{subfigure}
    \hfill
    \begin{subfigure}[b]{0.12\textwidth}
        \includegraphics[width=\textwidth]{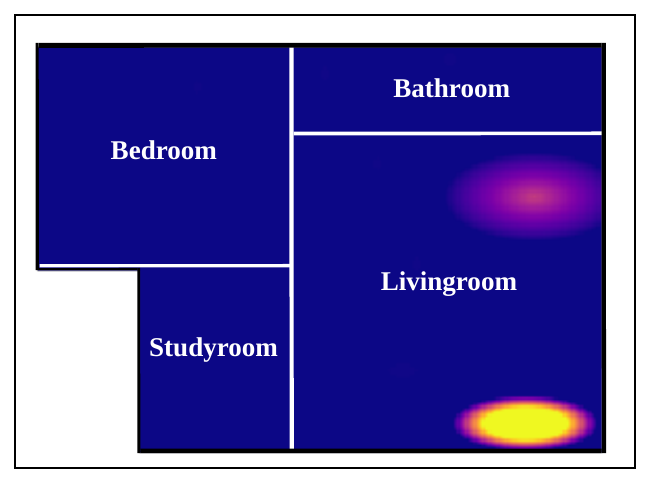}
    \end{subfigure}
    \hfill
    \begin{subfigure}[b]{0.12\textwidth}
        \includegraphics[width=\textwidth]{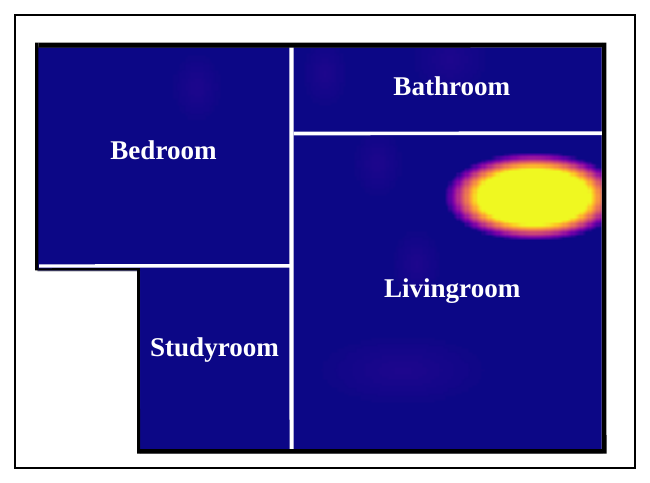}
        
    \end{subfigure}
    }
    \caption{The robot is tasked with searching for the phone, which is under the TV. The landmarks visited are: Start (Bedroom) $\xrightarrow{}$ Bed (Bedroom) $\xrightarrow{}$ Table (Studyroom) $\xrightarrow{}$ TV (Living). During this search, the robot finds the laptop on the studyroom table and the remote under the TV.}
    \label{fig:DBU}
\end{figure}

The motivation behind DBU is to improve performance in multi-object search. When assigned with multiple tasks, the robot uses its personalized beliefs for the first search. For the next search, the updated beliefs due to our DBU mechanism are used. This reduces the overall time and distance travelled during the search. The qualitative impact of DBU is shown in Fig. \ref{fig:DBU}. The robot is tasked with searching for the Phone but simultaneously keeps track of all other objects during the search. Fig. \ref{fig:DBU} shows the internal belief propagation of the robot for other objects (i.e., Cup, Remote, and Laptop) as it visits different landmarks. After the Phone is found, the robot returns to the start position. If the robot is then tasked with searching for either of the above items, it can search for them faster. Quantitative results for the impact of DBU are given in the ablation study (Sec. \ref{Sec: Ablation}). 

\vspace{-0.3cm}

\vspace{-0.3cm}

\section{Experimental Studies} \label{Sec: Experimental_studies}
We present the results of our experiments in this section. We study the influence of initial estimates, personalization, and adaptive Inferencing through the experiments. All experiments were conducted in the Living Lab at Tohoku University. The Living Lab is designed to accelerate research in service robots and human-robot interactions \cite{ravankar2023care}. The facility resembles a house to the extent possible.
We use the Turtlebot as our mobility platform and the ROS as our communication framework. 
We use YOLOv7, TEB local planner \cite{teb_planner} and Spatio-Temporal Voxel Layer \cite{Spatio-Temporal-Voxel-Layer} for navigation.% The \textit{Turtlebot} is equipped with an RPLidar S2 sensor and an Azure Kinect RGB-D sensor. In this section, we present the results of our experiments.

%\vspace{-0.4cm}

%\subsection{Experimental Setup}
%

\vspace{-0.4cm}

\subsection{Initial Estimate Comparison\label{Sec: Initial_Estimate}}
Initial estimate is an important aspect of any episode-based approach. With our framework (Sec. \ref{Sec: Personalization_Framework}), we compared the effectiveness of three different initial estimates in personalization. We sampled object locations from the multinomial distribution and the robot progressively personalized its belief over each episode from each initial estimate. The estimates are as follows:
\begin{itemize}
    \item Uniform estimate: All objects have equal probability at all landmarks. This estimate is void of any bias.
    \item In-situ estimate: The robot navigates through the environment for five episodes and builds an initial estimate based on the frequency of observations.
    \item Empirical estimate: We mine the Places365 and COCO datasets to obtain probabilistic relationships. 

\begingroup
\fontsize{9.5}{1.5}\selectfont
\begin{align}\label{eq:Empirical_Estimate}
    P(O|T) = \frac{\textbf{N}(O\cap T)+\lambda}{\textbf{N}(O})+\lambda n
\end{align}
\endgroup
\end{itemize}

Where, $n$ represents the total size of classes in the dataset, \textbf{N}(.) represents the tuples observed and $\lambda (=0.5)$ accounts for unobserved tuples and is set according to Jeffrey-Perk's law. 
%Performance comparison of different initial estimates can be found in Table \ref{tab:Initial_Estimate_Compare}.

The number of episodes required for termination for each object are given in Table \ref{tab:Initial_Estimate_Compare}. Five runs were conducted and the summary is presented below.
%The episode number is an average obtained after conducting five runs. 

\begingroup
\setlength{\tabcolsep}{3.5pt}
\begin{table}[h]
    \renewcommand{\arraystretch}{1.2}
    \centering
    \scriptsize
    \begin{tabular}{|c|c c c |c c c|c c c|}
        \hline
        & \multicolumn{9}{c|}{Initial Estimate} \\
        \cline{2-10}
        \multirow{3}{*}{Object} & \multicolumn{3}{c|}{Uniform} & \multicolumn{3}{c|}{In-Situ} & \multicolumn{3}{c|}{Empirical} \\
        \cline{2-10}
        & Min. & Max. & Avg. & Min. & Max. & Avg. & Min. & Max. & Avg. \\
        \hline
       
        Cup & 32&62 & \textbf{51} &33 & 99 & 59 & 32 & 92 & 57 \\
        Remote & 7 & 15 & 12 & 9 & 29 & 21 & 7 & 18 &\textbf{11} \\
        Book &23 &49 &39 &28 &91 &57 &14 &63 & \textbf{27} \\
        Laptop &22 &33 & \textbf{26} & 25 &56 &39 &25 &58 &41 \\
        Toothbrush &7 &7 &\textbf{7} &7 &7 &\textbf{7} &7 &7 &\textbf{7} \\
        Bottle &19 &33 & \textbf{25} &14 &58 &39 &34 &58 &48 \\
        Teddy Bear &11 &17 & 14 &10 &26 &16 &10 &23 & \textbf{13} \\
        Phone &23 &36 & \textbf{28} &24 &41 &33 &37 &63 &50 \\
        \hline
    \end{tabular}
    \captionsetup{belowskip=-5pt}
    \caption{Episodes required for different initial estimates.}
    \label{tab:Initial_Estimate_Compare}
\end{table}
\endgroup

%\vspace{-0.5cm}

\begin{figure}[h]
    \centering
    \begin{subfigure}[]{0.24\textwidth}
        \includegraphics[width=\textwidth]{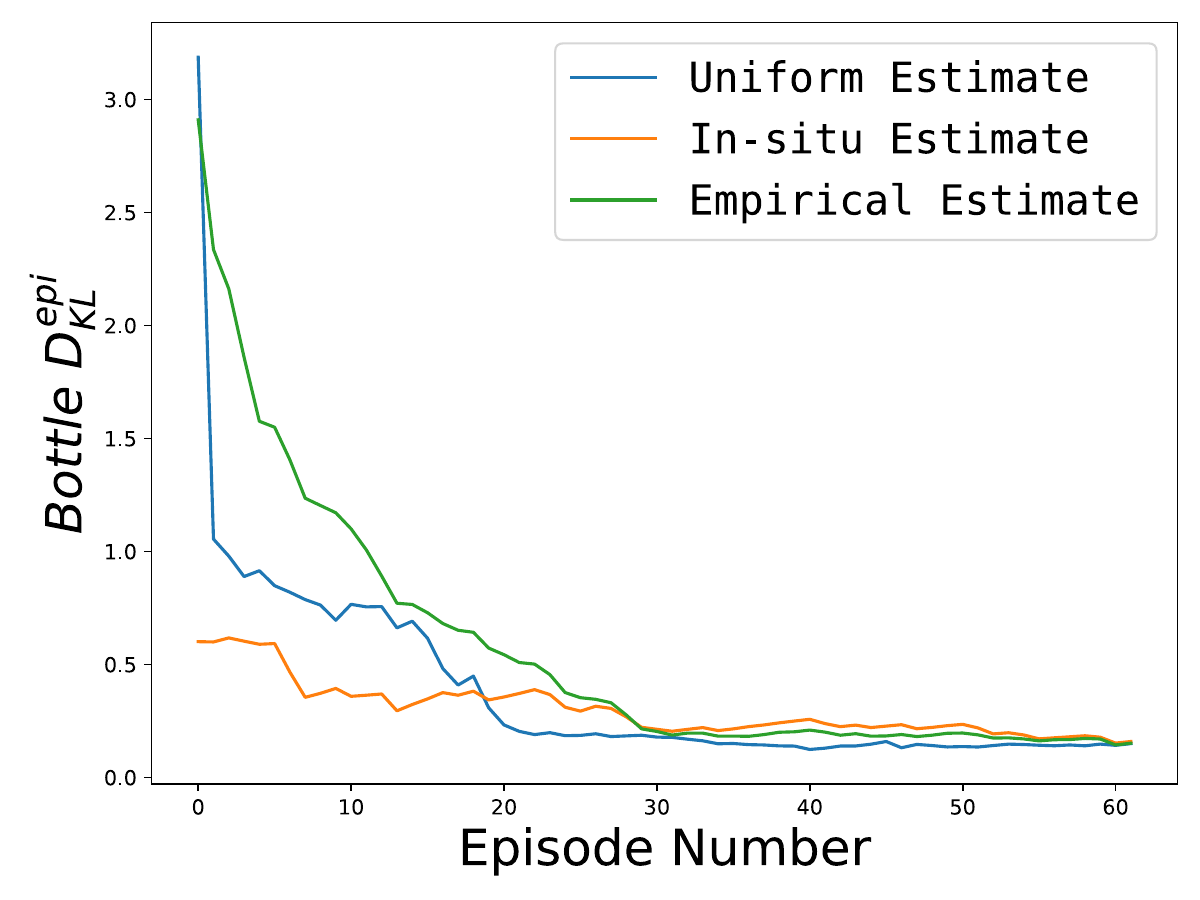}
        \captionsetup{justification=centering}
        \caption{Object: Bottle}
        \label{Fig:KL_Divergence_book} 
    \end{subfigure}
    \hfill
    \begin{subfigure}[]{0.24\textwidth}
        \includegraphics[width=\textwidth]{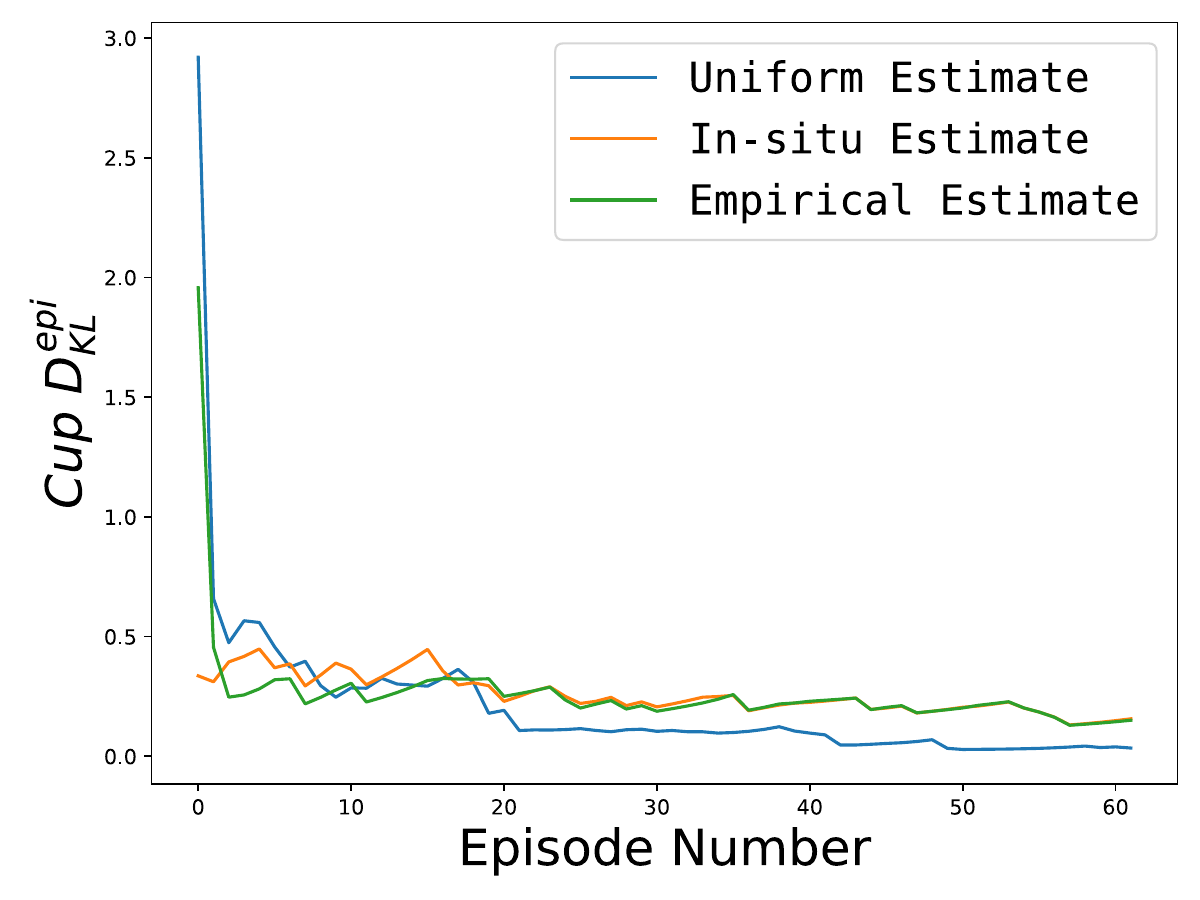}
        \captionsetup{justification=centering}
        \caption{Object: Cup}
        \label{Fig:KL_Divergence_cup}   
    \end{subfigure}
    
    \caption{KL Divergence for object-landmark-room relationships with respect to true values}.
    \label{fig:KL_Divergence}
\end{figure}

For each initial estimate, we also compute the KL Divergence statistic for object-landmark-room relationships for the episodes. This is shown in Fig. \ref{fig:KL_Divergence}. For an object, the KL Divergence with smoothing ($\epsilon=0.0001$) is computed as:

\begingroup
\fontsize{9.5}{1.5}\selectfont
\begin{align}
    D^{epi}_{KL}(P||P_{\text{true}}) =  \sum\limits_{\forall T}P(O|T)\log\left[ \frac{P(O|T)+\epsilon}{P_{\text{true}}(O|T)+\epsilon} \right]
\end{align}
\endgroup

Table \ref{tab:Initial_Estimate_Compare} and Fig. \ref{fig:KL_Divergence} compares the performance of all three estimates. The performance of the In-situ estimate is dependent on the multinomial sampling during the initial episodes. The empirical estimate is more accurate, but its performance declines if the true values deviate further from the estimate. Due to these reasons, from here on, we consider all probabilistic relationships based on the uniform initial estimate.

\vspace{-0.4cm}

\subsection{Ablation Study}\label{Sec: Ablation}
We consider personalization and DBU as separate modules for the ablation study. Subsequently, the baseline only consists of empirical estimate of priors (see Eq. \ref{eq:Empirical_Estimate}) along with our inferencing strategy. We perform the study on a multi-object search. For this, we task the robot to search for two objects sequentially. The second object is only revealed after the robot completes the first task. The distance for first task ($\text{D}_{1}$), cumulative distance ($\text{D}_{\text{T}}$), and total number of landmarks visited (Ldmk.) during the two tasks are shown in Table \ref{Tab:Ablation Study}. 

\begingroup
\setlength{\tabcolsep}{2pt}
\begin{table}[h]
    \renewcommand{\arraystretch}{1.2}
    \centering
    \scriptsize
    \begin{tabular}{|c|c c c|c c c|c c c|}
    \hline
    \multirow{2}{*}{Search Tasks}& \multicolumn{3}{c|}{A} & \multicolumn{3}{c|}{A + B} & \multicolumn{3}{c|}{A + B + C}  \\
    \cline{2-10}
    & $\text{D}_\text{1}\text{(m)}$ & $\text{D}_{\text{T}}\text{(m)}$  & Ldmk. & $\text{D}_\text{1}\text{(m)}$ & $\text{D}_{\text{T}}\text{(m)}$ & Ldmk. & $\text{D}_\text{1}\text{(m)}$ & $\text{D}_{\text{T}}\text{(m)}$ & Ldmk. \\
    \hline
    Phone+Laptop &21.14&46.61&15 & \textcolor{black}{17.71}&36.23&6  &\textcolor{black}{16.81}&21.73&4\\
    %Teddy+Book & & & &  31.73 & 6 & 177.2 & 14.90 & 3 & 83.76 \\
        Cup+Remote &\textcolor{black}{9.89}&\textcolor{black}{26.20}&\textcolor{black}{6} &\textcolor{black}{9.01}&26.44&5 &\textcolor{black}{9.13}&22.99&4 \\
    Phone+Book &\textcolor{black}{23.42}&\textcolor{black}{55.63}&\textcolor{black}{18} &\textcolor{black}{14.92}&33.92&6 &\textcolor{black}{15.21}&33.93&6 \\
    Laptop+Bottle &\textcolor{black}{28.89}&\textcolor{black}{41.67}&\textcolor{black}{11} &\textcolor{black}{8.82}&20.11&4 &\textcolor{black}{8.91}&14.64&3 \\
    Cup+Book &\textcolor{black}{11.7}&\textcolor{black}{48.69}&\textcolor{black}{14} &\textcolor{black}{6.43}&26.1&5 &\textcolor{black}{6.22}&24.75&4 \\ 
    \hline
    Average &\textcolor{black}{19.01}&\textcolor{black}{43.76}&\textcolor{black}{12.8} &\textcolor{black}{11.38}&29.69&5.4  & \textcolor{black}{\textbf{11.26}} & \textbf{21.64} & \textbf{4}\\
    \hline
    \multicolumn{10}{l}{\textcolor{black}{Module A: Baseline\ \ \ \ Module B: Personalization\ \ \ \ Module C: DBU}}
    \end{tabular}
    \captionsetup{justification=centering}
    \caption{Ablation study with personalization and DBU.}
    \label{Tab:Ablation Study}
\end{table}
\endgroup

The effects of including personalization and DBU for multi-object search are evident. We see that inclusion of personalization improved the $D_{\text{1}}$ metric by \textcolor{black}{68\%} and reduced the overall distance by \textcolor{black}{47\%} as compared to the Baseline. DBU doesn't affect the first task. Due to this, the average $D_{\text{1}}$ value remains the same compared to including personalization. However, the $D_{\text{T}}$ metric reduces by 37\% with the inclusion of DBU. %Overall the landmark visits are reduced by  } 

\begingroup
\setlength{\tabcolsep}{2.5pt}
\begin{table*}[t!]
    \renewcommand{\arraystretch}{1.15}
    \centering
    \scriptsize
    \begin{tabular}[t]{|c|c c c|c c c|c c c|c c c|c c c|c c c|c c c|}
        \hline
        %& \multicolumn{9}{c|}{Initial Estimate} \\
        %\cline{2-10}
            \multirow{2}{*}{Object} & \multicolumn{3}{c|}{PKS} & \multicolumn{3}{c|}{LTOS} & \multicolumn{3}{c|}{HSKOS} & \multicolumn{3}{c|}{HOS} & \multicolumn{3}{c|}{Ours (Ep=15)} & \multicolumn{3}{c|}{Ours (Ep=30)} & \multicolumn{3}{c|}{Ours (Ep=62)} \\
        \cline{2-22}\hspace{-2pt}
        & D(m) & Ldmk. & T(s) & D(m) & Ldmk. & T(s) & D(m) & Ldmk. & T(s) & D(m) & Ldmk.& T(s) & D(m) & Ldmk.& T(s) & D(m) & Ldmk. & T(s) & D(m) & Ldmk. & T(s)  \\
        \hline
       
        Cup &16.1&3&72.9 &11.01&4.3&60.5 &8.9&2.3&37.7 &7.1&-&43.6 &7.7&2&35.5 &7.5&2&35.9 &6.1&2&30.9\\
        Book &29.01&7.3&126.9 &19.3&6.3&101.9 &24.6&8&150.2 &11.1&-&60.1 &20.8&4&86.7 &17.03&3.7&81.9 &14.02&2.3&69.2 \\
        %Remote &13.3&3.3&59 &7.91&2.7&42.5 &9.65&2.7&47.2 &-&-&- &10.04&2.3&46.04 &9.16&2&42.42 &9.27&2&42.15  \\
        %Laptop &35.4&7&144.2 &16.9&6.3&91.2 &16.3&5.7&78.9 &-&-&- &14.6&3.3&50.6 &9.42&2&36.78 &9.47&2&36.39 \\
        Teddy &18.5&3.3&76.1 &11.1&3.7&68.4 &9.7&3.7&46.8 &6.3&-&24.7 &12.13&2&58.3 &7.2&2&27.9 &6.97&2&28.2  \\
        Bottle  &14.8&3&61.6 &13.3&5&71.6 &15.7&3.7&71.4 &10.1&-&61.3 &12.8&3&61.4 &9.4&2&48.7 &9.4&2&50.4 \\
        Phone &16.9&4&86.2 &17.7&6.3&88.8 &19.2&5.3&95.6 &10.5&-&57.2 &9.2&2.3&43.8 &10.01&2&41.4 &9.8&2&35.1 \\
        \hline
        Mean &19.1&4.1&84.7 &14.5&5.1&78.2 &15.6&4.6&80.3 &\textbf{8.2}&-&49.4 &12.7&2.7&57.1 &10.2&2.4&47.2 &9.3&\textbf{2.1}&\textbf{42.8} \\
        \hline
    \end{tabular}
    \captionsetup{belowskip=-10pt}
    \caption{Comparison of the overall framework with state of the art and baseline methods.} 
    \label{Tab:SOTA_Compare}
\end{table*}
\endgroup

\vspace{-0.3cm}

\subsection{Comparisons with State of the Art and Baselines} \label{subsec:SOTA_Comparison}
From the comparison, we intend to quantify the impact due to our two proposed novelties, viz. ontology personalization and adaptive inferencing. During the comparative experiments, the robot uses the same path planning algorithm and configuration of the navigation stack for maintaining consistency. The differences exist in the probabilistic associations and inferencing strategies which are the core value propositions of our approach. We compare against the following methods:

\begin{itemize}
    \item PKS: PKS is a probabilistic greedy approach where the next landmark to navigate to is the one with the highest probability of finding the object. This is a baseline method for benchmarking. %This strategy neither considers any long-horizon planning nor any belief updates.
    \item LTOS: The authors in \cite{Scene-graph-search} minimize the cost of different traversal routes according to the target. Each route is weighted (WPL) as follows:%use graph structures to represent the environment. Their inferencing is based on minimizing the cost of different traversal routes according to the target object. Each route is weighted (WPL) as follows:
    
    \begingroup
    \fontsize{9.5}{1.5}\selectfont
    \begin{align}
        \min WPL = \sum^{n}\limits_{i=0}\frac{Dist(L_{i},A)}{(1+\alpha*P(O|L_{i}))*2^{i}}
    \end{align}
    \endgroup
    
    \item HSKOS: 
    In \cite{Hierarchial_OS}, the authors consider the relative room size $R_{s}$, the distance between the agent and the room $Dist(R,A)$ and the probability of finding the object in the room $P(R|O)$ with the following Heuristic:
    
    \begingroup
    \fontsize{9.5}{1.5}\selectfont
    \begin{align}
        \argmaxA_R G(R)=\frac{P(R|O)}{R_{s}+Dist(R,A)}
    \end{align}
    \endgroup
    \item Human Operated Search (HOS): Humans are considered the gold standard for executing service robotic tasks. The human teleoperates the robot to search for the object. The human is given a preview tour of the environment.%without the presence of target objects. 
    \item Ours: We compare the \iffalse aforementioned \fi  SOTAs and baselines against our strategy. We consider three different levels of personalization. This corresponds to the beliefs after episode 15, 30, and 62. 
\end{itemize}

%the published methods. 

During the experiments, the robot (or operator) is tasked with a single object search, starting from the same location. We place the target object at its three most likely locations, which are defined as the locations with the highest probability according to the values visualized in Fig. \ref{Fig:True Values}. The distance (D) and time (T) to search are recorded. %The same conditions are considered in later parts (Fig. \ref{fig:Inf_Compare} and Fig. \ref{fig:Enhancement_Study}). We use the average distance and average number of landmarks visited as parameters for evaluation. 
Averages of the data recorded from our experiments are shown in Table \ref{Tab:SOTA_Compare}. %Thus a total of $1\cdot3\cdot5\cdot4 = 60$ experiments were conducted. 
%This set serves to prove the importance of personalization and our inferencing.

\begin{figure}[h]
    \centering
    \includegraphics[width=0.43\textwidth]{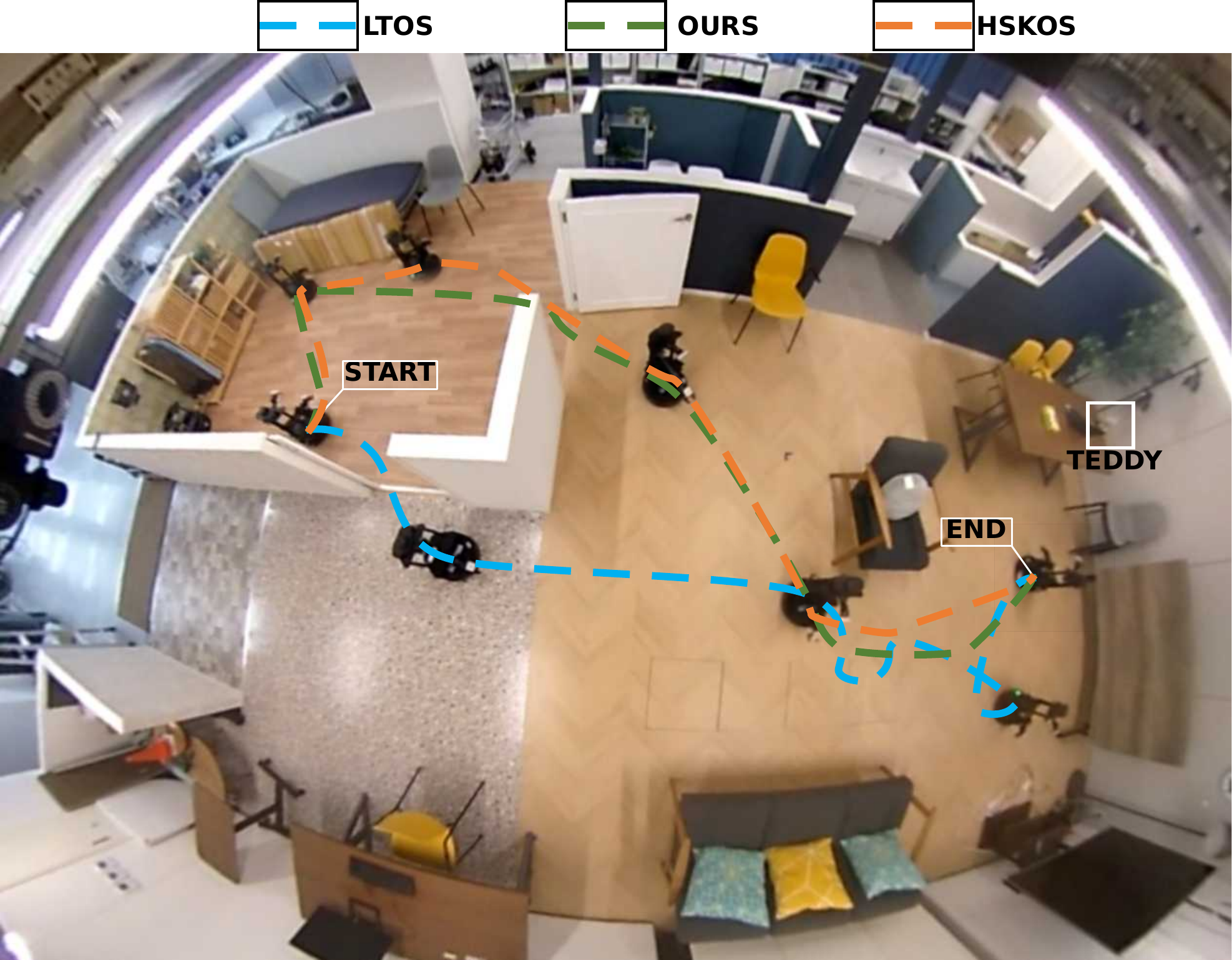}

    \captionsetup{justification=centering, skip=3pt}     
    \caption{Trajectories followed to search for Object: Teddy in the environment. PKS is omitted to improve clarity.}
    \label{fig:SOTA_Trajectory}
\end{figure}

The results in Table \ref{Tab:SOTA_Compare} show that our method outperforms both the SOTA as well as baseline and is close to human performance. LTOS has a better distance-to-landmark ratio since it is a long-horizon planner and HSKOS has fewer room transitions due to its hierarchical strategy. Despite this, our method outperforms HSKOS by around 33\% in terms of time and distance and LTOS by 41\% in time and 30\% in distance. Sample trajectories are shown in Fig. \ref{fig:SOTA_Trajectory} for reference.
%Despite this, both the methods do not perform as well as ours in terms of distance as well as landmarks visited.

\begin{figure}[h]
    \centering
    \begin{subfigure}[]{0.24\textwidth}
        \includegraphics[width=\textwidth]{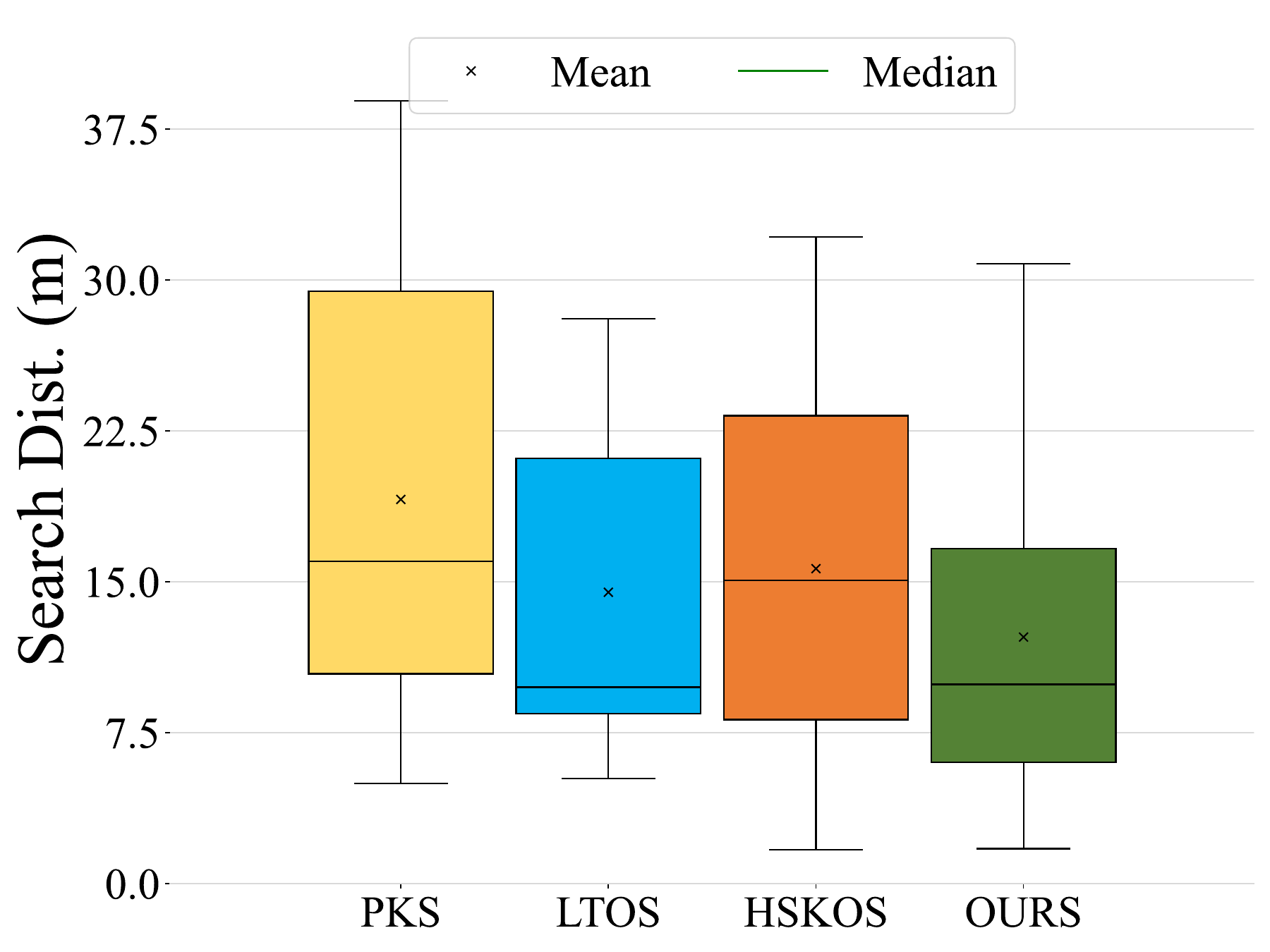}
        \captionsetup{aboveskip=0pt}
        \caption{Search distance}
        \label{fig:Inf_Dist}
    \end{subfigure}
    \hfill
    \begin{subfigure}[]{0.24\textwidth}
        \includegraphics[width=\textwidth]{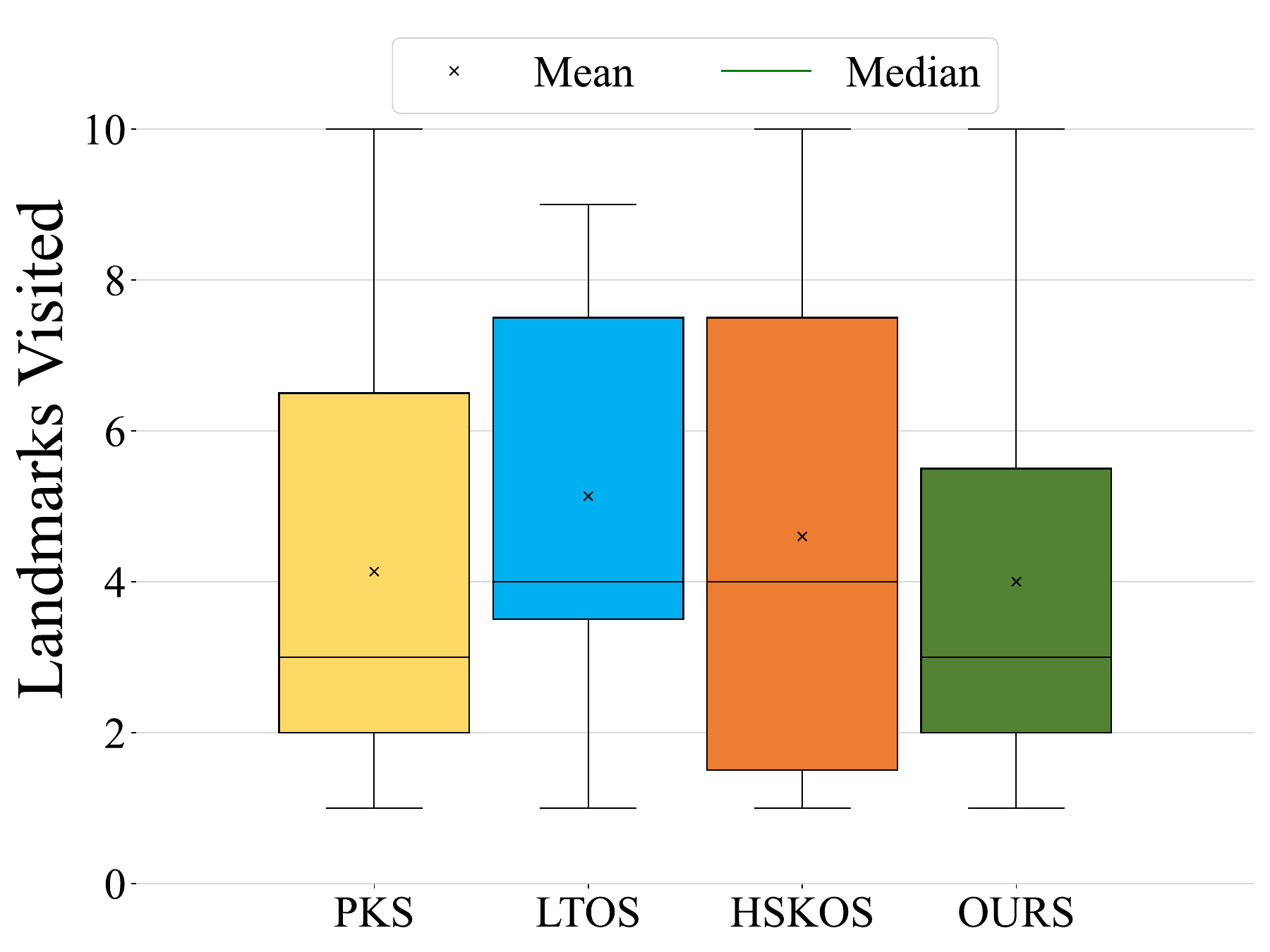}
        \captionsetup{aboveskip=0pt}
        \caption{Landmark visits}
        \label{fig:Inf_Ldmk}
    \end{subfigure}
    \captionsetup{justification=centering,aboveskip=3pt}
    \caption{Comparisons with respect to inferencing strategy.}
    \label{fig:Inf_Compare}
\end{figure}

We compared the frameworks as a whole in Table \ref{Tab:SOTA_Compare} above. Now, we compare the influence of our adaptive inferencing alone. We consider the strategies of PKS, LTOS and HSKOS for the comparison. PKS and HSKOS are greedy planners while LTOS is a long-horizon planner. However, all the methods have a fixed horizon unlike ours which adapts the horizon based on the ontology priors and distances. In this comparison, all methods use the empirical estimate for the ontology prior. Other conditions are same as for Table \ref{Tab:SOTA_Compare} and results are given in Fig. \ref{fig:Inf_Compare}.

\vspace{-0.2cm}

We see that only LTOS has comparable search distance (Fig. \ref{fig:Inf_Dist}) and only PKS has comparable landmark visits (Fig. \ref{fig:Inf_Ldmk}).
%Looking at Fig. \ref{fig:Inf_Dist}, we see that only LTOS has comparable performance. Considering Fig. \ref{fig:Inf_Ldmk}, only PKS has comparable performance. %our method has a better median distance value compared to others (10.2\% - 40.3\%). 
However, a nuanced look amplifies the improvements due to our approach. While PKS may have comparable landmark visits, the distance travelled is much greater ($\approx$61\%) since it does not consider distances while planning. On the contrary, LTOS may have comparable search distance but is worse ($\approx$33\%) on landmark visits  since it optimizes for worst-case scenarios. HSKOS has a more rounded performance but is worse on both the metrics. Additionally, HSKOS has a much larger inter-quartile range ($\approx$53\%)  for landmark visits than ours representing higher variability in performance.

\vspace{-0.4cm}

\subsection{Enhancement Study}
%While the comparative study against SOTAs demonstrates the superiority due to our approach,
We conduct an enhancement study to understand the standalone influence of personalization.
We claim that other SOTA methods can also benefit from personalization of ontologies. %To support this claim, we conduct an Enhancement Study.} 
In this context, we arm the SOTAs with our personalized relationships and use their respective planners. Other conditions are same as for Table \ref{Tab:SOTA_Compare} and the results are given below. 

\begin{figure}[h!]
    \centering
        \includegraphics[width=0.48\textwidth]{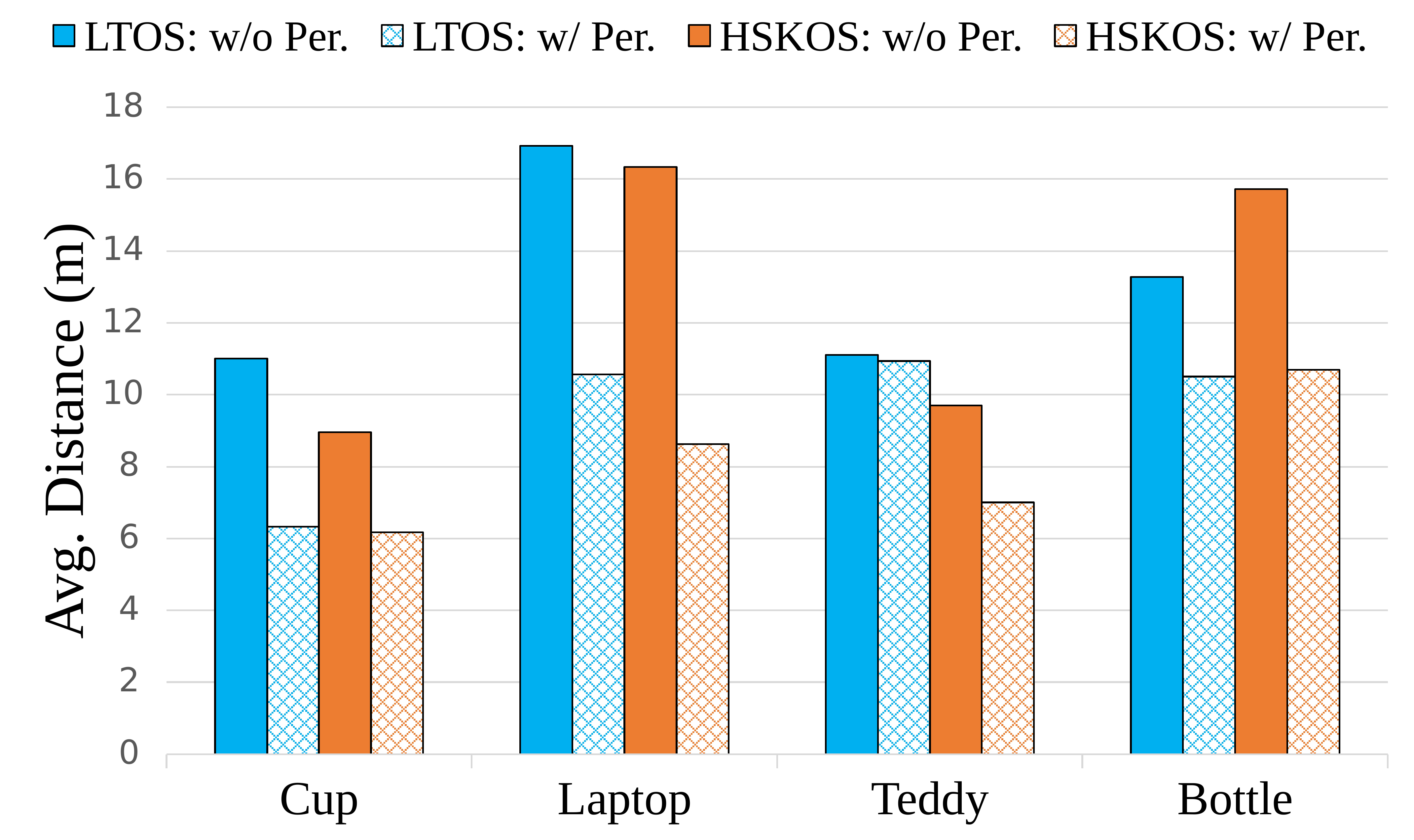}
    \captionsetup{justification=centering,belowskip=-5pt}
    \caption{Enhancement of SOTA with personalization.}
    \label{fig:Enhancement_Study}
\end{figure}

The performance of LTOS and HSKOS improved by around 25\%  and 35\% respectively, in terms of average distance. Correspondingly, HSKOS as well as LTOS required two less landmark visits on average. Thus, our personalization framework can be used as a catalyst for improving the performance of other SOTA methods. 

\vspace{-0.2cm}

\section{Conclusions and Discussions}
We presented a novel approach for the \textit{personalization} of ontologies of \textit{object-landmark-room} relationships. Our methodology drastically improves efficiency of object search as the robot learns more about the environment. We tested our framework with three different initial estimates and determined that a uniform initial estimate fares better in personalization as compared to others. We also proposed an \textit{adaptive inferencing} strategy with a formal approach for deciding the lookahead window. The strategy adapts the search horizon based on the ontology prior. The inferencing strategy included dynamic belief updates for better execution of multi-object search tasks. The ablation study confirmed that the inclusion of ontology personalization and DBU improved the performance compared to the baseline for multi-object search. Our framework was able to outperform multiple SOTAs in real environments. %The proposed framework was able to outperform the SOTAs on multiple parameters. %such as average landmarks visited, search time and distance to search. 
Our adaptive inferencing strategy fared better than the SOTA strategies for the same priors. We showed that an adaptive horizon provides better performance than fixed horizon during planning. Additionally, the use of our personalization framework as a catalyst improved the performance of SOTAs. We thus developed a framework that not only provides better results than SOTAs but is also able to bolster their performance.

An important assumption in our work, however, is the fixed position of landmarks. In its truest sense, long-term would mean that the locations of landmarks can change, or even new landmarks can be added or removed. Secondly, the object categories are limited, whereas an open-vocabulary object search would be more desirable for long-term applications.
%Apart from this, the entire indoor environment may change as well (for eg. when shifting to a new apartment).
Thus, in the future, we would like to develop meta-learning strategies that are robust to such dynamic scenarios.

\vspace{-0.2cm}

%%% My Library can't get the style exactly the same even with the instructions%%%
\bibliographystyle{IEEEtran}
\bibliography{references2.bib} %change to your file name (with suffix .bib)

\begin{thebibliography}{10}
\providecommand{\url}[1]{#1}
\csname url@samestyle\endcsname
\providecommand{\newblock}{\relax}
\providecommand{\bibinfo}[2]{#2}
\providecommand{\BIBentrySTDinterwordspacing}{\spaceskip=0pt\relax}
\providecommand{\BIBentryALTinterwordstretchfactor}{4}
\providecommand{\BIBentryALTinterwordspacing}{\spaceskip=\fontdimen2\font plus
\BIBentryALTinterwordstretchfactor\fontdimen3\font minus \fontdimen4\font\relax}
\providecommand{\BIBforeignlanguage}[2]{{%
\expandafter\ifx\csname l@#1\endcsname\relax
\typeout{** WARNING: IEEEtran.bst: No hyphenation pattern has been}%
\typeout{** loaded for the language `#1'. Using the pattern for}%
\typeout{** the default language instead.}%
\else
\language=\csname l@#1\endcsname
\fi
#2}}
\providecommand{\BIBdecl}{\relax}
\BIBdecl

\bibitem{vimantic}
D.~Fernandez-Chaves, J.-R. Ruiz-Sarmiento, N.~Petkov, and J.~Gonzalez-Jimenez, ``Vimantic, a distributed robotic architecture for semantic mapping in indoor environments,'' \emph{Knowledge-Based Systems}, vol. 232, p. 107440, 2021.

\bibitem{My_House_My_Rules}
I.~Kapelyukh and E.~Johns, ``My house, my rules: Learning tidying preferences with graph neural networks,'' in \emph{Proceedings of the 5th Conference on Robot Learning}.\hskip 1em plus 0.5em minus 0.4em\relax PMLR, 2022, pp. 740--749.

\bibitem{TidyBot}
J.~Wu, R.~Antonova, A.~Kan, M.~Lepert, A.~Zeng, S.~Song, J.~Bohg, S.~Rusinkiewicz, and T.~Funkhouser, ``Tidybot: personalized robot assistance with large language models,'' \emph{Autonomous Robots}, vol.~47, no.~8, p. 1087–1102, Nov. 2023.

\bibitem{IshiiRisk}
I.~Sena, A.~Chikhalikar, A.~A. Ravankar, J.~V.~S. Luces, and Y.~Hirata, ``Context-aware risk estimation in home environments: A probabilistic framework for service robots,'' in \emph{2025 IEEE International Conference on Robot and Human Interactive Communication (ROMAN)}, 2025.

\bibitem{pmlr-v205-patel23a}
M.~Patel and S.~Chernova, ``Proactive robot assistance via spatio-temporal object modeling,'' in \emph{Proceedings of The 6th Conference on Robot Learning}, vol. 205, 2023, pp. 881--891.

\bibitem{patel2023predicting}
M.~Patel, A.~G. Prakash, and S.~Chernova, ``Predicting routine object usage for proactive robot assistance,'' in \emph{7th Annual Conference on Robot Learning}, 2023.

\bibitem{Panoptic_TSDFs}
L.~Schmid, J.~Delmerico, J.~L. Schönberger, J.~Nieto, M.~Pollefeys, R.~Siegwart, and C.~Cadena, ``Panoptic multi-tsdfs: a flexible representation for online multi-resolution volumetric mapping and long-term dynamic scene consistency,'' in \emph{2022 International Conference on Robotics and Automation (ICRA)}, 2022, pp. 8018--8024.

\bibitem{rosinol2021kimera}
A.~Rosinol, A.~Violette, M.~Abate, N.~Hughes, Y.~Chang, J.~Shi, A.~Gupta, and L.~Carlone, ``Kimera: From slam to spatial perception with 3d dynamic scene graphs,'' \emph{The International Journal of Robotics Research}, vol.~40, no. 12-14, pp. 1510--1546, 2021.

\bibitem{habitatchallenge2022}
K.~Yadav, S.~K. Ramakrishnan, J.~Turner, A.~Gokaslan, O.~Maksymets, R.~Jain, R.~Ramrakhya, A.~X. Chang, A.~Clegg, M.~Savva, E.~Undersander, D.~S. Chaplot, and D.~Batra, ``Habitat challenge 2022,'' 2022.

\bibitem{BRM_OS}
C.~Wang, J.~Cheng, J.~Wang, X.~Li, and M.~Q.-H. Meng, ``Efficient object search with belief road map using mobile robot,'' \emph{IEEE Robotics and Automation Letters}, vol.~3, no.~4, pp. 3081--3088, 2018.

\bibitem{Multi_scale_POMDPs}
L.~Holzherr, J.~F{\"o}rster, M.~Breyer, J.~Nieto, R.~Siegwart, and J.~J. Chung, ``Efficient multi-scale pomdps for robotic object search and delivery,'' in \emph{2021 IEEE International Conference on Robotics and Automation (ICRA)}.\hskip 1em plus 0.5em minus 0.4em\relax IEEE, 2021, pp. 6585--6591.

\bibitem{Multi-res-POMDP}
K.~Zheng, Y.~Sung, G.~Konidaris, and S.~Tellex, ``Multi-resolution pomdp planning for multi-object search in 3d,'' in \emph{2021 IEEE/RSJ International Conference on Intelligent Robots and Systems (IROS)}, 2021, pp. 2022--2029.

\bibitem{Opt-Corr-Obj-Search}
K.~Zheng, R.~Chitnis, Y.~Sung, G.~Konidaris, and S.~Tellex, ``Towards optimal correlational object search,'' in \emph{2022 International Conference on Robotics and Automation (ICRA)}, 2022, pp. 7313--7319.

\bibitem{Viewpoint-selection}
A.~C. Hernandez, E.~Derner, C.~Gomez, R.~Barber, and R.~Babuška, ``Efficient object search through probability-based viewpoint selection,'' in \emph{2020 IEEE/RSJ International Conference on Intelligent Robots and Systems (IROS)}, 2020, pp. 6172--6179.

\bibitem{Semantic-Temporal}
M.~Mantelli, F.~M. Noori, D.~Pittol, R.~Maffei, J.~Torresen, and M.~Kolberg, ``Semantic temporal object search system based on heat maps,'' \emph{Journal of Intelligent \& Robotic Systems}, vol. 106, no.~4, p.~69, 12 2022.

\bibitem{Semantic-Grounding}
Y.~Zhang, G.~Tian, X.~Shao, M.~Zhang, and S.~Liu, ``Semantic grounding for long-term autonomy of mobile robots toward dynamic object search in home environments,'' \emph{IEEE Transactions on Industrial Electronics}, vol.~70, no.~2, pp. 1655--1665, 2023.

\bibitem{Hierarchial_OS}
M.~Zhang, G.~Tian, Y.~Cui, Y.~Zhang, and Z.~Xia, ``Hierarchical semantic knowledge-based object search method for household robots,'' \emph{IEEE Transactions on Emerging Topics in Computational Intelligence}, vol.~8, no.~1, pp. 930--941, 2024.

\bibitem{Scene-graph-search}
F.~Zhou, H.~Liu, H.~Zhao, and L.~Liang, ``Long-term object search using incremental scene graph updating,'' \emph{Robotica}, vol.~41, no.~3, 2023.

\bibitem{Semantic_driven_OS}
I.~Remy, A.~Gupta, and K.~Leung, ``Semantically-driven object search using partially observed 3d scene graphs,'' in \emph{NeurIPS 2023 Foundation Models for Decision Making Workshop}, 2023.

\bibitem{AkashOpenVocab}
A.~Chikhalikar, A.~A. Ravankar, J.~V.~S. Luces, and Y.~Hirata, ``Open vocabulary object search utilizing large language models and fuzzy inferencing,'' in \emph{2025 IEEE/SICE International Symposium on System Integration (SII)}, 2025, pp. 345--351.

\bibitem{wang2019semantic}
C.~Wang, J.~Cheng, W.~Chi, T.~Yan, and M.~Q.-H. Meng, ``Semantic-aware informative path planning for efficient object search using mobile robot,'' \emph{IEEE Transactions on Systems, Man, and Cybernetics: Systems}, vol.~51, no.~8, pp. 5230--5243, 2021.

\bibitem{multimodal_object_search}
S.~Hasegawa, A.~Taniguchi, Y.~Hagiwara, L.~El~Hafi, and T.~Taniguchi, ``Integrating probabilistic logic and multimodal spatial concepts for efficient robotic object search in home environments,'' \emph{SICE Journal of Control, Measurement, and System Integration}, vol.~16, no.~1, pp. 400--422, 2023.

\bibitem{yolov7}
C.-Y. Wang, A.~Bochkovskiy, and H.-Y.~M. Liao, ``Yolov7: Trainable bag-of-freebies sets new state-of-the-art for real-time object detectors,'' \emph{arXiv preprint arXiv:2207.02696}, 2022.

\bibitem{kuhn1955hungarian}
H.~W. Kuhn, ``{The Hungarian method for the assignment problem},'' \emph{Naval research logistics quarterly}, vol.~2, no. 1-2, pp. 83--97, 1955.

\bibitem{akashSII}
A.~Chikhalikar, A.~A. Ravankar, J.~V.~S. Luces, S.~A. Tafrishi, and Y.~Hirata, ``An object-oriented navigation strategy for service robots leveraging semantic information,'' in \emph{2023 IEEE/SICE International Symposium on System Integration (SII)}, 2023, pp. 1--6.

\bibitem{ipa_room_segmentation}
R.~Bormann, F.~Jordan, W.~Li, J.~Hampp, and M.~Hägele, ``Room segmentation: Survey, implementation, and analysis,'' in \emph{2016 IEEE International Conference on Robotics and Automation (ICRA)}, 2016, pp. 1019--1026.

\bibitem{Places365}
B.~Zhou, A.~Lapedriza, A.~Khosla, A.~Oliva, and A.~Torralba, ``Places: A 10 million image database for scene recognition,'' \emph{IEEE Transactions on Pattern Analysis and Machine Intelligence}, 2017.

\bibitem{Complexity-search-path}
K.~Trummel and J.~Weisinger, ``The complexity of the optimal searcher path problem,'' \emph{Operations Research}, vol.~34, no.~2, pp. 324--327, 1986.

\bibitem{Decision_making_framework}
T.~H. Chung and J.~W. Burdick, ``A decision-making framework for control strategies in probabilistic search,'' in \emph{2007 IEEE International Conference on Robotics and Automation}, 2007, pp. 4386--4393.

\bibitem{ravankar2023care}
A.~A. Ravankar, S.~A. Tafrishi, J.~V.~S. Luces, F.~Seto, and Y.~Hirata, ``Care: Cooperation of ai-robot enablers to create a vibrant society,'' \emph{IEEE Robotics and Automation Magazine}, 2023.

\bibitem{teb_planner}
C.~Rösmann, F.~Hoffmann, and T.~Bertram, ``{Integrated online trajectory planning and optimization in distinctive topologies},'' \emph{Robotics and Autonomous Systems}, vol.~88, pp. 142--153, 2017.

\bibitem{Spatio-Temporal-Voxel-Layer}
S.~Macenski, D.~Tsai, and M.~Feinberg, ``Spatio-temporal voxel layer: A view on robot perception for the dynamic world,'' \emph{International Journal of Advanced Robotic Systems}, vol.~17, no.~2, 2020.

\end{thebibliography}

%%% Example Library %%%

%\bibliographystyle{IEEETran}
%\bibliography{myRef} %change to your file name (with suffix .bib)

\end{document}